\documentclass{article}

\usepackage[final]{neurips_2025}

\usepackage[utf8]{inputenc} 
\usepackage[T1]{fontenc}    
\usepackage{hyperref}       
\usepackage{url}            
\usepackage{booktabs}       
\usepackage{amsfonts}       
\usepackage{nicefrac}       
\usepackage{microtype}      
\usepackage{xcolor}         
\usepackage{tablefootnote}
\usepackage[dvipsnames,svgnames]{xcolor}
\usepackage{array}
\usepackage{diagbox}

\usepackage[normalem]{ulem}
\usepackage{bm}
\usepackage{enumitem}

\usepackage{amsmath}
\usepackage{amssymb}
\usepackage{mathtools}
\usepackage{amsthm}
\usepackage{wrapfig}
\usepackage{multirow}
\usepackage{subcaption}
\theoremstyle{plain}

\theoremstyle{definition}

\theoremstyle{remark}

\newcommand{\n}{TAMI}
\newcommand{\ote}{original TE}
\newcommand{\lte}{LTE}
\newcommand{\trc}{LHA}

\usepackage[most]{tcolorbox}
\definecolor{propbg}{HTML}{F2F2F9}
\definecolor{propfr}{HTML}{00007B}
\newtcolorbox{propositionbox}{
  enhanced,
  boxrule=0pt,frame hidden,
  borderline west = {2pt}{0pt}{propfr},
  colback=propbg,
  sharp corners
}

\definecolor{darkgreen}{RGB}{0, 0, 205}

\title{\n{}: Taming Heterogeneity in Temporal Interactions for Temporal Graph Link Prediction}

\author{%
  Zhongyi Yu$^{1}$,
  \quad Jianqiu Wu$^{1}$,
  \quad Zhenghao Wu$^{2}$,
  \quad Shuhan Zhong$^{3}$,\\ 
  \quad \textbf{Weifeng Su}$^{1,5}$,
  \quad \textbf{Chul-Ho Lee}$^{4}$,
  \quad \textbf{Weipeng Zhuo}$^{1,5}$\thanks{Corresponding author.}\\  
  \\
  $^1$Beijing Normal-Hong Kong Baptist University,
  $^2$University College Dublin\\
  $^3$The Hong Kong University of Science and Technology, 
  $^4$Texas State University\\
  $^5$Guangdong Provincial / Zhuhai Key Laboratory of IRADS, China
  \\
  \texttt{\{zhongyiyu,jianqiuwu,wfsu,weipengzhuo\}@uic.edu.cn,} \\
  \texttt{zhenghao.wu@ucdconnect.ie,} \texttt{szhongaj@ust.hk,} \texttt{chulho.lee@txstate.edu}
}

\begin{document}

\maketitle

\begin{abstract}
Temporal graph link prediction aims to predict future interactions between nodes in a graph based on their historical interactions, which are encoded in node embeddings. We observe that heterogeneity naturally appears in temporal interactions, e.g., a few node pairs can make most interaction events, and interaction events happen at varying intervals. This leads to the problems of ineffective temporal information encoding and forgetting of past interactions for a pair of nodes that interact intermittently for their link prediction. Existing methods, however, do not consider such heterogeneity in their learning process, and thus their learned temporal node embeddings are less effective, especially when predicting the links for infrequently interacting node pairs. To cope with the heterogeneity, we propose a novel framework called \n{}, which contains two effective components, namely log time encoding function (\lte{}) and link history aggregation (\trc{}). \lte{} better encodes the temporal information through transforming interaction intervals into more balanced ones, and \trc{} prevents the historical interactions for each target node pair from being forgotten. State-of-the-art temporal graph neural networks can be seamlessly and readily integrated into \n{} to improve their effectiveness. Experiment results on 13 classic datasets and three newest temporal graph benchmark (TGB) datasets show that \n{} consistently improves the link prediction performance of the underlying models in both transductive and inductive settings. Our code is available at \url{https://github.com/Alleinx/TAMI_temporal_graph}.
\end{abstract}

\section{Introduction}
\label{sec/intro}
Temporal link prediction is a fundamental task that forecasts future interactions between two nodes on continuous-time temporal graphs (CTTGs). This is important in a variety of real-world scenarios with dynamic graph topologies changing over time, such as social networks~\cite{social_2,social_1}, user-item interaction systems~\cite{JODIE,user_item2,new_user_item_1,new_user_item_2}, traffic networks~\cite{traffic1,new_traffic_1,new_traffic_2}, and physical systems~\cite{physics,new_physics_1}. The temporal link prediction task involves two major steps. The first step is to compute the temporal embedding of each node by aggregating information from its own historical interactions with neighboring nodes. For each target node pair for link prediction, the second step is to take their embeddings into a link predictor (typically an MLP) to estimate the probability of having a link between them.

\begin{wrapfigure}{r}{0.47\textwidth} 
    \centering    
    \includegraphics[width=.45\columnwidth]{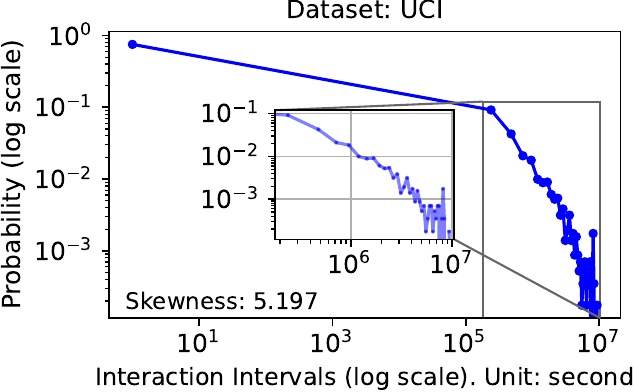}    
    \caption{The interaction intervals between any pair of nodes on the UCI dataset follow a power-law distribution and have a high (positive) skewness value, meaning that the intervals are highly right-skewed.}    
    \label{fig:uci_frequency_distribution}
\end{wrapfigure}

Heterogeneity arises in temporal interactions. For example, the number of interactions for each node pair varies substantially. The time between two consecutive interaction events can also be quite different. To see this, in Figure~\ref{fig:uci_frequency_distribution}, we plot the distribution of interaction intervals between any pair of nodes on the UCI dataset, showing that it follows a power-law distribution. In other words, most interactions are frequent ones, while the \emph{infrequent} ones (very large interaction intervals) still exist with a non-negligible probability. We also compute and report the Fisher's skewness in Figure~\ref{fig:uci_frequency_distribution}, indicating that the distribution is positively skewed or right-skewed. Here the skewness $\Gamma$ of a random variable $X$ is defined as $\Gamma = \mathbb{E} \left[ (X - \mu)^3/\sigma^3 \right]$, where $\mu$ and $\sigma$ denote the mean and standard deviation of $X$, respectively.

The heterogeneity in temporal interactions poses two primary challenges to existing methods for temporal link prediction~\cite{tgat,graphmixer,DyGFormer}. First, they use sinusoidal functions as their time encoding functions to encode the temporal difference between the target time $\tau$ and each interaction time $t_i$. We observe that the frequency parameters of the time encoding functions are \emph{harder to learn} when the distribution of the temporal difference is highly skewed, which is the common case as seen from the distribution of interaction intervals. Second, they learn the embedding of each node based only on its recent interaction events. While the embedding encodes timely information, it can be predominantly influenced by a few neighbors having frequent interactions. As a result, the temporal link prediction can be inaccurate for a target node pair having \emph{infrequent} interactions over time. For example, it is natural to predict that a couple would eat hamburgers if it is one of their recent favorites (frequent interactions). However, it makes more sense to predict that turkey will be on the table (infrequent interactions) if Thanksgiving is coming.

To address the above challenges, we propose \n{}, a novel framework that \textbf{tam}es the heterogeneity in temporal \textbf{i}nteractions to improve the performance of temporal link prediction. \n{} mainly contains two novel components, namely log time encoding function (\lte{}) and link history aggregation (\trc{}). We propose to use a logarithmic transformation in \lte{} to rescale the temporal difference before it is taken into the time encoding function, so that the frequency parameters of the time encoding function can be easier (or quicker) to learn. In addition, we develop \trc{} to preserve the information of the most recent $k$ interactions for each target node pair for link prediction, regardless of their interaction frequency. In other words, the historical interactions of each node pair are explicitly used for their link prediction, no matter when they happened.

Our key contributions can be summarized as follows:

\begin{itemize}[leftmargin=*, partopsep=0pt, itemsep=6pt]

    \item \textbf{First study on the heterogeneity in temporal interactions.} To the best of our knowledge, this is the first work to identify the presence of heterogeneity in temporal interactions of CTTGs and investigate its impact on the performance of temporal link prediction.
    
    \item \textbf{\n{}: a novel framework to cope with the heterogeneity in temporal interactions.} We propose \n{}, which effectively handles the heterogeneity in temporal interactions. \n{} contains two novel modules, namely \lte{} and \trc{}. \lte{} rescales the temporal differences using a logarithmic transformation while \trc{} is specifically designed to preserve historical interactions for each target node pair for link prediction. Existing temporal graph neural networks can be seamlessly and readily integrated into \n{}.

    \item \textbf{Extensive evaluation on open datasets}. We validate the effectiveness of \n{} on 13 classic temporal graph datasets covering different fields with comprehensive temporal scales as well as three newest ones from temporal graph benchmark (TGB). Results show that \n{} consistently and substantially improves the link prediction accuracy and training efficiency of the underlying graph neural networks, with up to 87.05\% improvement in link prediction accuracy and 76.7\% reduction in total training time.      
\end{itemize}

\section{Related Work}
\label{sec:related_work}
\textbf{Time Encoding Functions in Link Prediction.} Most prior methods~\cite{tgat,graphmixer,DyGFormer,CAWN,TNCN,TCL,TGN,rebuttal_linear_tpe} use a common approach for time encoding, which is based on either periodic time encoding functions introduced in~\cite{original_tpe} or their simplified ones. To predict the existence of a link between two nodes at the target time, the temporal difference between the target time and the time of each historical interaction needs to be calculated. The time differences are then mapped to time encoding vectors using a periodic function such as sinusoidal function. However, this time encoding pipeline overlooks the effect of the skewness in the time differences. We observe that it takes longer than needed to train a model, and it can also degrade the performance of link prediction, especially when predicting the links for infrequently interacting node pairs. In this work, we propose \lte{} to effectively alleviate the skewness in the time differences via a simple yet effective logarithmic transformation, leading to better training efficiency and improved link prediction accuracy.

\textbf{Temporal Neighborhood Aggregation in Link Prediction.} In learning temporal node embeddings using information aggregation, there are typically two types of temporal graph neural networks (TGNNs), which are random walk-based TGNNs and temporal neighbor-based TGNNs. Random walk-based TGNNs~\cite{CAWN,random_walk_1,rebuttal_random_walk} first generate multiple causal, anonymous walks for each node and then aggregate information from these walks to obtain the final node embeddings. Since frequent interactions predominantly appear as edges in the graph, it may be less likely to sample infrequent interactions in each walk. In addition, temporal neighbor-based (or graph convolution-based) TGNNs~\cite{DyRep,JODIE,TGN,memory_disttgl,memory_chen_rnn,related_TN-based_2,tgat,TCL,graphmixer,DyGFormer,2024_iclr_frequency_dyg,memory_APAN,new_temporal_1,new_temporal_2,new_temporal_3,rebuttal_neighbor} compute temporal node embeddings by aggregating node or edge features from the temporal neighbors of each target node. For example, recurrent networks or temporal point processes~\cite{DyRep} are leveraged in~\cite{JODIE,memory_chen_rnn,DyRep} to store the states of historical interactions. DyGFormer~\cite{DyGFormer} computes node embeddings by aggregating an extensive number of single-hop temporal neighbors using an attention mechanism. GraphMixer~\cite{graphmixer} proposes to use MLP-Mixer~\cite{MLPMixer} and neighbor mean-pooling to compute node embeddings. However, these methods mainly rely on the recent interactions of a node to compute its embedding, which are again dominated by frequent interactions with a limited number of its neighbors. Thus, they can be less effective in predicting the links for infrequently interacting node pairs. In contrast, we propose \trc{} to effectively address this problem by explicitly capturing the historical interactions between each target node pair for link prediction in the modeling process.

\section{\n{} Design}
\label{sec/model}
\begin{figure*}[t]
\begin{center}
\centerline{\includegraphics[width=\textwidth]{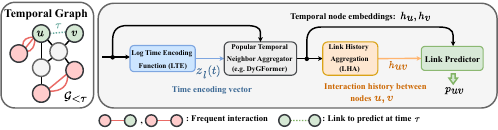}}
\caption{The \n{} framework.}
\label{fig:BALI-framework}
\end{center}
\vskip -0.4in
\end{figure*}
\subsection{Problem Definition}

A CTTG over a time interval $[0,T]$ is characterized by $\mathcal{G}_T \!=\! (\mathcal{V}_T, \mathcal{E}_T)$, which is a sequence of chronologically ordered interaction events between nodes up to time $T$ (inclusive), where $\mathcal{V}_T$ is the set of nodes and $\mathcal{E}_T$ is the set of temporal edges up to $T$. Here, a temporal edge $e_{uv}^t \in \mathcal{E}_T$ between two nodes $u \!\in\! \mathcal{V}_T$ and $v \!\in\! \mathcal{V}_T$ represents an interaction event between $u$ and $v$ at time $t \!<\! T$. Note that multiple edges can exist between two nodes in the temporal graph.

Given two target nodes $u$ and $v$, the target time $\tau$, and all historical interaction events in the graph up to $\tau$ (exclusive) that are characterized by $\mathcal{G}_{<\tau}$, our task is to predict how likely nodes $u$ and $v$ will interact with each other at time $\tau$, as illustrated in Figure~\ref{fig:BALI-framework}. In other words, it is to estimate the probability $p_{uv}$ of having a link between $u$ and $v$ at time $\tau$. To tackle this problem, there are two common major steps involved in most TGNNs~\cite{tgat,graphmixer,DyGFormer,TNCN,t_rewriting}. First, they compute the temporal embeddings of nodes $u$ and $v$ by aggregating information from their recent neighbors who have recently had interactions with them. Second, the learned embeddings are then fed into a link predictor, e.g., an MLP with a sigmoid activation function, to estimate the probability $p_{uv}$.

\subsection{\lte{}: Log Time Encoding Function}
\label{sec:model_log-te}

In most TGNNs~\cite{tgat,graphmixer,DyGFormer,rebuttal_random_walk,rebuttal_neighbor}, each historical interaction at time $t < \tau$ (or its corresponding temporal edge in $\mathcal{G}_{<\tau}$) is associated with a time encoding vector $\bm{z}(t)$, which is given by
\begin{equation}
\label{eq:original_tpe}
    \bm{z}(t) = \cos(\Delta t \times \bm{\omega}), 
\end{equation}
with $\Delta t = \tau - t$, and learnable frequency parameters $\bm{\omega} \!=\! \{\alpha^{-(i-1) / \beta}\}_{i=1}^{d_T}$, where $d_T$ is the dimension of the time encoding vector, and the values of $\alpha$ and $\beta$ are initialized as $\alpha = \beta = \sqrt{d_T}$. This time encoding function first maps $\Delta t$ to a monotonically decreasing vector $\Delta t \times \bm{\omega}$ such that the values are within $(0, \Delta t]$ and then projects them to $[-1, 1]$ using the cosine function. For the sake of clarity, we refer to this type of time encoding as \ote{}.

\begin{figure*}[t!]
    \centering
   
        \begin{subfigure}[t]{0.48\textwidth}
        \centering
        \includegraphics[width=\textwidth]{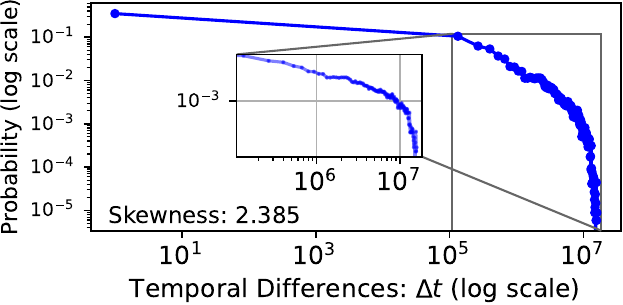}
        \caption{The distribution of temporal difference $\Delta t$ on the UCI dataset is highly right-skewed.}
        \label{fig:skew_enron_presentation}
    \end{subfigure}%
    \hspace{0.02\textwidth}
    \begin{subfigure}[t]{0.48\textwidth}
        \centering
        \includegraphics[width=\textwidth]{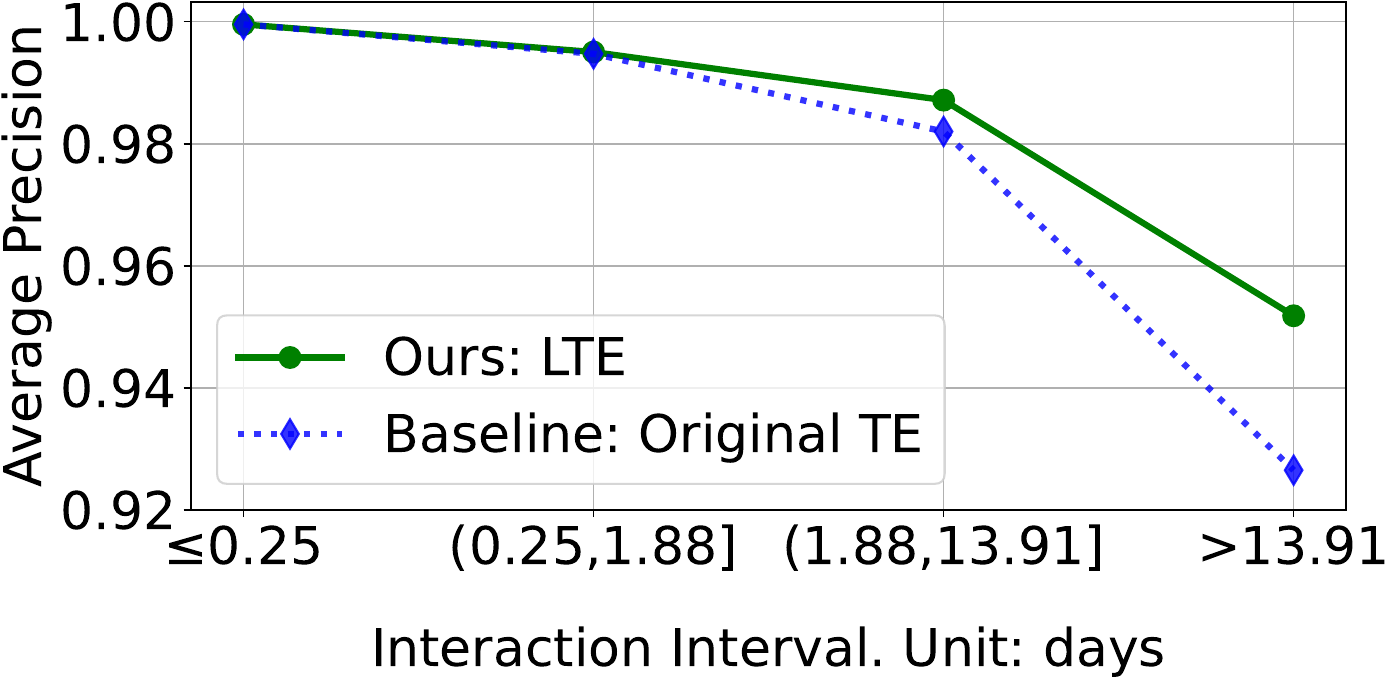}
        \caption{\lte{} effectively mitigates the skewness and improves the model (GraphMixer~\cite{graphmixer}) performance.}
        \label{fig:tpe_frequency_ap}
    \end{subfigure}
    \caption{(a) Distribution of temporal difference $\Delta t$ and (b) model performance on the UCI dataset.}
\end{figure*}

We observe that a given node can interact with various other nodes at different frequencies. Thus, their interaction intervals can vary significantly. As shown in Figure~\ref{fig:uci_frequency_distribution}, the interaction intervals between nodes are highly right-skewed on the UCI dataset. Such right skewness also appears in the distribution of temporal differences $\Delta t$, as can be seen from Figure~\ref{fig:skew_enron_presentation}. TGNNs trained on these skewed inputs using \ote{} learn better on the interactions that frequently appear but struggle with the ones that seldom occur. As a result, they are ineffective when making a link prediction for a pair of nodes that interact rarely or whose interaction intervals are long, as shown in Figure~\ref{fig:tpe_frequency_ap}, where we group testing node pairs for link prediction according to their average interaction intervals and report the average prediction accuracy for each group of node pairs.

To address this challenge, we propose \lte{}, a simple yet effective time encoding function, which rescales the value of $\Delta t$ via a logarithmic transformation so that it follows a more balanced distribution. With \lte{}, a large variance in $\Delta t$ no longer leads to a large discrepancy, making it easier to learn the frequency parameters $\bm{\omega}$. The time encoding function in \lte{} is formally defined as
\begin{equation}
\label{eq:log-te}
    \bm{z}_l(t) = \cos(\Delta t_l \times \bm{\omega}), \text{ with~}
    \Delta t_l = \ln(1 + \Delta t).
\end{equation}
To see how the logarithmic transformation mitigates the skewness in the distribution, we consider $\Delta t$ to follow a Pareto distribution, which is a power-law distribution. We have the following:

\begin{propositionbox}
\label{prop_box}    
    \textbf{Proposition 1.} \textit{Suppose $\Delta t$ follows a Pareto distribution with the shape parameter $\alpha > 3$, whose skewness is always greater than 2. \lte{} reduces the skewness to 2.} 
\end{propositionbox}
The proof is provided in Appendix~\ref{sec:proof_prop_1}.

In practice, $\Delta t$ may not strictly follow a Pareto distribution. Nonetheless, \lte{} can still effectively mitigate its skewness, thereby improving the performance of the underlying model. As shown in Figure~\ref{fig:tpe_frequency_ap}, \lte{} improves the model performance by reducing the skewness of $\Delta t$ on the UCI dataset from $2.385$ to $-1.14$. Please refer to Table~\ref{tab:skewness_of_all_datasets_appendix} in Appendix~\ref{appendix:dataset_skewness} for the values of skewness with and without \lte{} on different datasets.

Once the time encoding vector is obtained, the next step in \n{} is to compute the temporal node embeddings. For this purpose, any existing TGNN~\cite{DyRep,JODIE,TGN,tgat,TCL,graphmixer,DyGFormer,CAWN} can be adopted. In temporal neighbor-based methods~\cite{DyGFormer,graphmixer,tgat,JODIE,TGN}, the temporal  embedding $\bm{h}_u \!\in\! \mathbb{R}^d$ of node $u$ is generally computed based on the $m$ most recent interactions of $u$ with its temporal neighbors. Specifically, let $\mathcal{N}_u$ be the set of the $m$ recent neighbors of $u$, $\bm{x}_j$ be the initial embedding of node $j$, and $\bm{x}_{uj}$ be the initial embedding of temporal edge $e_{uj}$, and $\bm{z}_l(t_j)$ be the time encoding vector for the interaction event with $j$ at time $t_j$. Then, the embedding $\bm{h}_u$ is obtained as follows:
\begin{equation}
    \bm{h}_u = \mathrm{AGGREGATE} \left (\{[\bm{x}_j; \bm{x}_{uj}; \bm{z}_l(t_j)]\}_{j \in \mathcal{N}_u} \right),
\end{equation}
where $\mathrm{AGGREGATE}(\cdot)$ is an aggregation function and $[;]$ is the concatenation operation. Similarly, random-walk based methods~\cite{CAWN,random_walk_1} obtain the embedding $\bm{h}_u$ by aggregating the information from nodes that appear in random walks starting from $u$. Both classes of methods can be readily integrated into \n{}, as shall be demonstrated in Section~\ref{sec:portability_main_text}.

\subsection{\trc{}: Link History Aggregation}
\label{sec:model_trc}

\begin{wrapfigure}{r}{0.51\textwidth} 
    \centering    
    \includegraphics[width=.49\columnwidth]{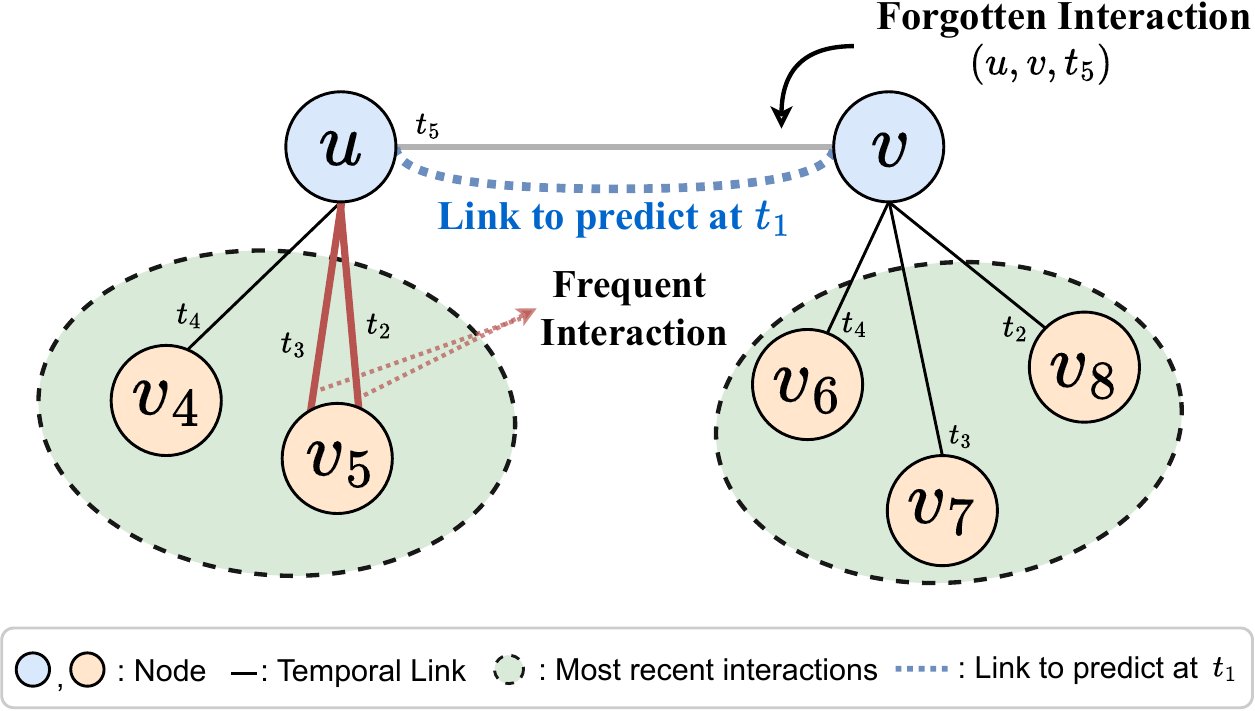}
    \caption{When predicting the future link between $u$ and $v$ at $t_1$, their historical interaction at $t_5$ is no longer retained in their latest node embeddings.}    
    \label{fig:k-hop-forget}
\end{wrapfigure}

\begin{figure*}[t]
\begin{center}
\centerline{\includegraphics[width=1\textwidth]{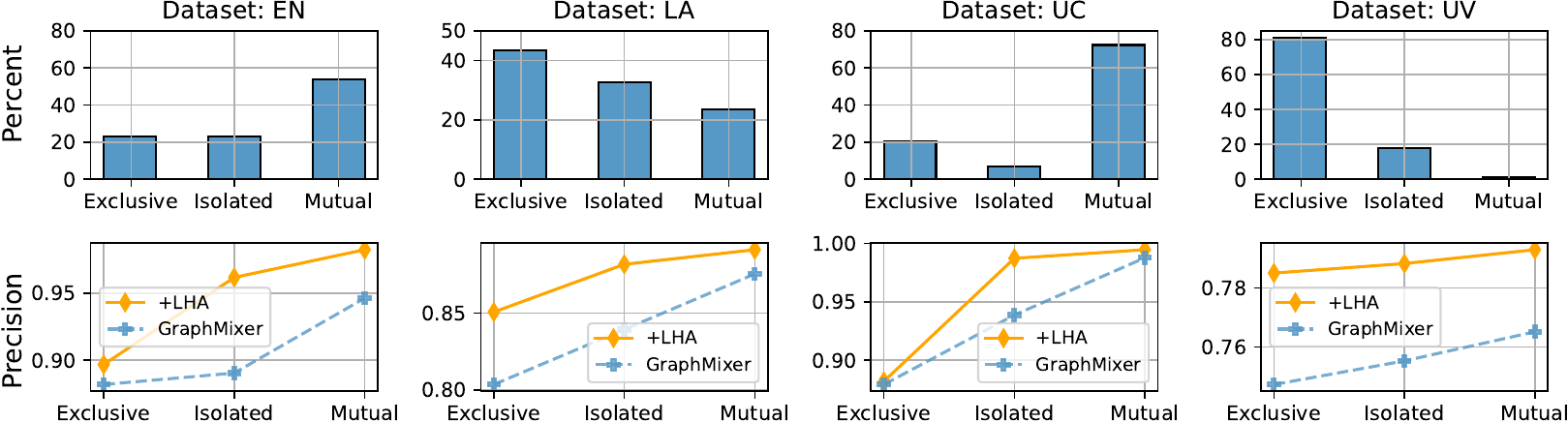}}
\caption{The performance of GraphMixer and its improvement with \trc{}. ``Exclusive'' means that neither of two nodes appears in the other's $m$ recent interactions. ``Isolated'' indicates that only one node appears in the other's $m$ recent interactions. ``Mutual'' means that both nodes appear.}
\label{fig:trc_direct_match}
\end{center}
\vskip -0.4in
\end{figure*}

As mentioned above, it is common practice that the temporal embedding of a node $u$ is based only on its most recent $m$ interactions or the recent ones that appear in random walks from $u$. However, this can be problematic for predicting a link between two nodes $u$ and $v$, especially when they do not appear in their mutual nearest neighbors. In that case, their historical interactions are forgotten in updating their temporal embeddings, thereby degrading the link prediction accuracy. See Figure~\ref{fig:k-hop-forget} for an illustration. We empirically observe that this is indeed the case, as shown in Figure~\ref{fig:trc_direct_match}. There is a non-negligible (possibly significant) portion of node pairs that could have forgotten their mutual interaction history for temporal link prediction, and their link prediction accuracy is the worst.

To resolve this problem, we propose a novel light-weight module called link history aggregation (\trc{}). Its core idea is to preserve the most recent $k$ interactions for a target node pair and leverage this historical information, along with their temporal node embeddings, to predict a link between the target node pair. Let $e_{uv}^{t_1}, e_{uv}^{t_2}, \dots, e_{uv}^{t_k}$ denote the most recent $k$ interactions between nodes $u$ and $v$ that happen at times $t_1$, $t_{2}, \ldots, t_k$ before time $\tau$, respectively, where $t_k < t_{k-1} < \cdots < t_1 < \tau$. The $i$-th historical interaction $e_{uv}^{t_i}$ is associated with a $d_r$-dimensional historical edge embedding $\bm{r}_{uv}^{t_i} \in \mathbb{R}^{d_r}$, which encodes the information of the interaction. We use $M_{uv}(\tau) = \{\bm{r}_{uv}^{t_1}, \bm{r}_{uv}^{t_2}, ... , \bm{r}_{uv}^{t_k}\}$ to indicate the historical edge embeddings of the most recent $k$ interactions before time $\tau$. Then, the edge embedding vector $\bm{r}_{uv}^{\tau}$ for link prediction at time $\tau$ is defined as a weighted sum of the current node embeddings and the previous edge embedding as:
\begin{equation}
\label{eq:trc_context_vector}
    \bm{r}_{uv}^{\tau} = \gamma \times \bm{c}_{uv} + (1 - \gamma) \times \bm{r}_{uv}^{t_1},
\end{equation}
where $\bm{c}_{uv} = \text{MLP}( [\bm{h}_u \, ; \, \bm{h}_v] ) \in \mathbb{R}^{d_r}$ encodes the current states of nodes $u$ and $v$, with $\bm{h}_u$ and $\bm{h}_v$ being the temporal node embeddings of nodes $u$ and $v$, respectively, and $\bm{r}_{uv}^{t_1}$ is the most recent edge embedding in $M_{uv}(\tau)$. Here the hyperparameter $\gamma \in [0, 1]$ controls the `forgetting' rate of historical interactions. For example, if $\gamma = 1$, the entire interaction history is discarded in computing $\bm{r}_{uv}^{\tau}$. To bootstrap the link prediction between $u$ and $v$ which do not have interaction history, i.e. $M_{uv}(\tau) = \emptyset$, we set $\bm{r}_{uv}^{t_1} = \bm{0}$.

To predict a link between nodes $u$ and $v$ at time $\tau$ as the last step of \n{}, we first aggregate all the historical link embeddings in $M_{uv}(\tau)$ into a single vector $\bm{h}_{uv}$, which summarizes the most recent $k$ interaction histories between $u$ and $v$. This can be written as
\begin{equation}
    \bm{h}_{uv} = \mathrm{AGGREGATE} \, (\bm{r}_{uv}^{t_1}, \bm{r}_{uv}^{t_2}, ... , \bm{r}_{uv}^{t_k}) \in \mathbb{R}^{d_r},
\end{equation}
where $\mathrm{AGGREGATE}(\cdot)$ denotes an aggregation function. While various options can be considered (such as sum, mean, or attention), we employ the \textbf{most-recent} aggregator. It utilizes the edge embedding of the most recent historical interaction to update $\bm{h}_{uv}$, i.e., $\bm{h}_{uv} = \bm{r}_{uv}^{t_1}$. Finally, the link probability $p_{uv}$ is computed using an MLP with the sigmoid activation function, i.e.,
\begin{equation}
    p_{uv} = \text{MLP} \big ( [\bm{h}_u \, ; \, \bm{h}_v \, ; \, \bm{h}_{uv}] \big ).
\end{equation}
Whenever an interaction between nodes $u$ and $v$ occurs, their corresponding history set $M_{uv}(\cdot)$ needs to be updated. Suppose $u$ and $v$ are connected at time $\tau$. A new edge embedding vector $\bm{r}_{uv}^{\tau}$, computed using Equation~\eqref{eq:trc_context_vector}, is added to $M_{uv}(\tau)$ for the next link prediction. If the original size of $M_{uv}(\tau)$ is equal to $k$, we first remove the oldest historical edge embedding from $M_{uv}(\tau)$ and then insert $\bm{r}_{uv}^{\tau}$. In other words, we keep the most recent $k$ interactions only. 

\textbf{Remarks.} As shown in Figure~\ref{fig:trc_direct_match} and shall be demonstrated in the subsequent section, \trc{} improves the performance of the underlying TGNN, with up to 25.33\% improvement in link prediction accuracy. \trc{} also has high efficiency in terms of GPU memory usage. Note that $\bm{h}_{uv}$ is the only additional embedding used in computing the link probability $p_{uv}$, and we use the \textit{most-recent} aggregator to obtain $\bm{h}_{uv} = \bm{r}_{uv}^{t_1}$. Also, the historical edge embedding $\bm{r}_{uv}^{t_1}$ is updated based on its previous one, as in Equation~\eqref{eq:trc_context_vector}. Thus, letting $N$ be the number of target node pairs for prediction in the dataset, the total space complexity of \trc{} is $O(N d_r)$, where $d_r$ is the dimension of each historical edge embedding. Since mini-batch computation can be employed, it is unnecessary to load the edge embeddings for all target node pairs in GPU memory. Instead, they can be stored in CPU's system memory and loaded dynamically as required. Therefore, the additional GPU memory overhead introduced by \trc{} is $O(b d_r)$, where $b$ denotes the mini-batch size, and $b \ll N$. Furthermore, with a slight increase in complexity, \trc{} also speeds up the convergence of the underlying model in training, with up to 76.7\% reduction in total training time, as shall be demonstrated shortly. 

\section{Experiments}
\label{sec/experiments}

\subsection{Experimental Settings}
\label{sec:exp_settings}
We conduct experiments on 13 classic open datasets~\cite{edgebank}: Can. Parl. (\textbf{CP}), Contact (\textbf{CO}), Enron (\textbf{EN}), Flights (\textbf{FL}), Lastfm (\textbf{LA}), Mooc (\textbf{MO}), Reddit (\textbf{RE}), Social Evo (\textbf{SE}), Uci (\textbf{UC}), UN Trade (\textbf{UT}), UN Vote (\textbf{UV}), US Legis (\textbf{US}), and Wikipedia (\textbf{WK}). These datasets cover various domains and their details are provided in Section~\ref{appendix:dataset_description}. We integrate two state-of-the-art TGNNs, namely GraphMixer~\cite{graphmixer} and DyGFormer~\cite{DyGFormer}, into our \n{} framework. More specifically, we use their temporal neighbor aggregation function in the Temporal Neighbor Aggregator module of \n{}. We compare the performance with their vanilla counterparts, as well as seven other state-of-the-art TGNNs for CTTGs, including JODIE~\cite{JODIE}, DyRep~\cite{DyRep}, TGAT~\cite{tgat}, TGN~\cite{TGN}, CAWN~\cite{CAWN}, Edgebank~\cite{edgebank}, and TCL~\cite{TCL}. Descriptions of the baselines are provided in Section~\ref{appendix:baseline_descriptions}.

We evaluate the link prediction performance of TGNNs under two settings: (1) the transductive setting where future links are predicted between nodes observed during training, and (2) the inductive setting where predictions are made for nodes unseen during training. Following~\cite{edgebank,DyGFormer}, we chronologically split each dataset into 70\%/15\%/15\% for training/validation/testing, and adopt the average precision (AP) score as the evaluation metric. We present the implementation details in Section~\ref{appendix:implementation_details} and baseline configurations in Section~\ref{appendix:config_baselines}. Unless otherwise specified, we below present the major results for the transductive setting only and report additional results for the transductive and inductive settings in Section~\ref{sec:appendix_transductive_results} and Section~\ref{appendix:inductive_link_prediction}, respectively. For the generation of negative links, we follow the negative sampling strategies in~\cite{edgebank}, where random negative sampling is used for training and random, historical, and inductive negative samplings are applied during evaluation, with random negative sampling as the default choice.

\subsection{Main Results}
\label{sec:main_results}

\newcolumntype{X}{>{\fontsize{7.7}{9}\selectfont}c}
\setlength{\tabcolsep}{1.5pt}
\begin{table*}[t]
\caption{AP for transductive link prediction {under three different negative sampling strategies (NSS)}. Imp. (\%) denotes the percentage of \emph{improvement}. The first and the second best performers are marked in \textbf{bold} and \underline{underlined}, respectively. Standard deviations over five runs are reported in Table~\ref{tab:std_main_table}.}

\label{tab:main_result}

\centering
\scriptsize
\begin{tabular}{X|X|XXXXXXXXXXXXX}
\toprule
\textbf{NSS} & \textbf{Methods} & \textbf{CP} & \textbf{CO} & \textbf{EN} & \textbf{FL} & \textbf{LA} & \textbf{MO} & \textbf{RE} & \textbf{SE} & \textbf{UC} & \textbf{UT} & \textbf{UV} & \textbf{US} & \textbf{WK} \\
\midrule

\multirow{14}{*}{rnd} & JODIE      & 69.26 & 95.31 & 84.77 & 95.60 & 70.85 & 80.23 & 98.31 & 89.89 & 89.43 & 64.94 & \underline{63.91} & 75.05 & 96.50 \\

& DyRep      & 66.54 & 95.98 & 82.38 & 95.29 & 71.92 & 81.97 & 98.22 & 88.87 & 65.14 & 63.21 & 62.81 & \underline{75.34} & 94.86 \\

& TGAT       & 70.73 & 96.28 & 71.12 & 94.03 & 73.42 & 85.84 & 98.52 & 93.16 & 79.63 & 61.47 & 52.21 & 68.52 & 96.94 \\

& TGN        & 70.88 & 96.89 & 86.53 & 97.95 & 77.07 & \textbf{89.15} & 98.63 & 93.57 & 92.34 & 65.03 & \textbf{65.72} & \textbf{75.99} & 98.45 \\

& CAWN       & 69.82 & 90.26 & 89.56 & 98.51 & 86.99 & 80.15 & 99.11 & 84.96 & 95.18 & \underline{65.39} & 52.84 & 70.58 & 98.76 \\

& EdgeBank   & 64.55 & 92.58 & 83.53 & 89.35 & 79.29 & 57.97 & 94.86 & 74.95 & 76.20 & 60.41 & 58.49 & 58.39 & 90.37 \\

& TCL        & 68.67 & 92.44 & 79.70 & 91.23 & 67.27 & 82.38 & 97.53 & 93.13 & 89.57 & 62.21 & 51.90 & 69.59 & 96.47 \\

& GraphMixer & 75.90 & 91.94 & 82.26 & 90.98 & 75.56 & 82.83 & 97.33 & 93.34 & 93.38 & 62.61 & 52.20 & 71.55 & 97.23 \\

& DyGFormer & \underline{97.91} & \underline{98.31} & \underline{92.46} & \underline{98.92} & \underline{93.01} & {87.66} & \underline{99.22} & \underline{94.66} & 95.66 & 65.07 & 55.88 & 70.44 & \underline{99.02} \\

\cmidrule{2-15}
& \multicolumn{13}{l}{\textbf{with \n{}}} \\
\cmidrule{2-15}

& {GraphMixer} & 78.38 & 95.26 & 90.97 & 96.75 & 88.13 & 83.53 & 98.84 & 93.41 & \underline{96.20} & 62.98 & 57.74 & 71.57 & 98.89 \\

& Imp. (\%) & {\color{darkgreen}3.27\%} & {\color{darkgreen}3.61\%} & {\color{darkgreen}10.59\%} & {\color{darkgreen}6.34\%} & {\color{darkgreen}16.64\%} & {\color{darkgreen}0.85\%} & {\color{darkgreen}1.56\%} & {\color{darkgreen}0.07\%} & {\color{darkgreen}3.02\%} & {\color{darkgreen}0.59\%} & {\color{darkgreen}10.61\%} & {\color{darkgreen}0.03\%} & {\color{darkgreen}1.71\%} \\

& {DyGFormer} & \textbf{98.67} & \textbf{98.70} & \textbf{92.66} & \textbf{98.94} & \textbf{94.03} & \underline{88.49} & \textbf{99.29} & \textbf{94.74} & \textbf{96.72} & \textbf{66.39} & 56.02 & {71.40} & \textbf{99.25} \\ 

& Imp. (\%) & {\color{darkgreen}0.78\%} & {\color{darkgreen}0.40\%} & {\color{darkgreen}0.22\%} & {\color{darkgreen}0.02\%} & {\color{darkgreen}1.10\%} & {\color{darkgreen}0.95\%} & {\color{darkgreen}0.07\%} & {\color{darkgreen}0.08\%} & {\color{darkgreen}1.11\%} & {\color{darkgreen}2.03\%} & {\color{darkgreen}0.25\%} & {\color{darkgreen}1.36\%} & {\color{darkgreen}0.23\%} \\ 

\midrule

\multirow{7}{*}{hist} & GraphMixer& 74.34 & 93.29 & 77.98 & 71.47 & 72.47 & 77.77 & 78.44 & 94.93 & 84.11 & 57.05 & 51.20 & 81.65 & {90.90} \\
& DyGFormer & {97.00} & {97.57} & 75.63 & 66.59 & {81.57} & 85.85 & 81.57 & {97.38} & 82.17 & 64.41 & 60.84 & 85.30 & 82.23 \\

\cmidrule{2-15}
& \multicolumn{13}{l}{\textbf{with \n{}}} \\
\cmidrule{2-15}

& GraphMixer & 78.81 & 93.30 & {81.68} & {73.01} & 80.23 & 83.61 & {82.56} & 96.80 & {87.69} & {69.74} & 70.90 & 84.56 & {90.97} \\
& Imp. (\%)       & {\color{darkgreen}6.01\%} & {\color{darkgreen}0.01\%} & {\color{darkgreen}4.74\%} & {\color{darkgreen}2.15\%} & {\color{darkgreen}10.71\%} & {\color{darkgreen}7.51\%} & {\color{darkgreen}5.25\%} & {\color{darkgreen}1.97\%} & {\color{darkgreen}4.26\%} & {\color{darkgreen}22.24\%} & {\color{darkgreen}38.48\%} & {\color{darkgreen}3.56\%} & {\color{darkgreen}0.08\%} \\

& DyGFormer  & {98.96} & {97.72} & {81.02} & 67.77 & {83.40} & {86.26} & {85.18} & {97.56} & {85.89} & 65.16 & {81.72} & {86.10} & 82.38 \\
& Imp. (\%)       & {\color{darkgreen}2.02\%} & {\color{darkgreen}0.15\%} & {\color{darkgreen}7.13\%} & {\color{darkgreen}1.77\%} & {\color{darkgreen}2.24\%} & {\color{darkgreen}0.48\%} & {\color{darkgreen}4.43\%} & {\color{darkgreen}0.18\%} & {\color{darkgreen}4.53\%} & {\color{darkgreen}1.16\%} & {\color{darkgreen}34.32\%} & {\color{darkgreen}0.94\%} & {\color{darkgreen}0.18\%} \\

\midrule

\multirow{7}{*}{ind} & GraphMixer & 69.48 & 90.87 & 75.01 & 74.87 & 68.12 & 74.26 & 85.26 & 94.72 & 80.10 & 60.15 & 51.60 & 79.63 & {88.59} \\

& DyGFormer & {95.44} & 94.75 & 77.41 & 70.92 & 73.97 & {81.24} & 91.11 & {97.68} & 72.25 & 55.79 & 51.91 & 81.25 & 78.29 \\

\cmidrule{2-15}
& \multicolumn{13}{l}{\textbf{with \n{}}} \\
\cmidrule{2-15}

& GraphMixer & 70.94 & {96.12} & {88.95} & {93.64} & {91.06} & 79.82 & {96.19} & 96.09 & {84.12} & {87.73} & {79.53} & {83.31} & {93.89} \\

& Imp. (\%) & {\color{darkgreen}2.10\%} & {\color{darkgreen}5.78\%} & {\color{darkgreen}18.58\%} & {\color{darkgreen}25.07\%} & {\color{darkgreen}33.68\%} & {\color{darkgreen}7.49\%} & {\color{darkgreen}12.82\%} & {\color{darkgreen}1.45\%} & {\color{darkgreen}5.02\%} & {\color{darkgreen}45.85\%} & {\color{darkgreen}54.13\%} & {\color{darkgreen}4.62\%} & {\color{darkgreen}5.98\%} \\

& DyGFormer & {97.25} & {98.47} & {86.23} & 75.55 & 74.03 & {92.39} & {94.37} & {97.76} & {80.13} & 68.01 & {78.19} & {81.31} & 78.96 \\

& Imp. (\%) & {\color{darkgreen}1.90\%} & {\color{darkgreen}3.93\%} & {\color{darkgreen}11.39\%} & {\color{darkgreen}6.53\%} & {\color{darkgreen}0.08\%} & {\color{darkgreen}13.72\%} & {\color{darkgreen}3.58\%} & {\color{darkgreen}0.08\%} & {\color{darkgreen}10.91\%} & {\color{darkgreen}21.90\%} & {\color{darkgreen}50.63\%} & {\color{darkgreen}0.07\%} & {\color{darkgreen}0.86\%} \\

\bottomrule
\end{tabular}
\vskip -0.1in
\end{table*}
\setlength{\tabcolsep}{6pt}

\noindent\textbf{Performance under different negative sampling strategies}. As shown in Table~\ref{tab:main_result}, \n{} substantially improves the performance of the integrated models under all three negative sampling strategies. In particular, for the random negative sampling strategy (rnd), the performance of both GraphMixer and DyGFormer improves on all 13 datasets, with improvement up to 16.64\%. This is because \lte{} balances the skewness in temporal differences during the time encoding process, {especially for nodes that interact rarely or whose interaction intervals are long}, while \trc{} prevents historical interactions between the target node pair from being forgotten in predicting their future links, improving the link prediction accuracy. In addition, we show in Section~\ref{sec:appendix_inductive_main} that \n{} remains effective in the inductive setting, consistently improving the performance of the integrated models. These results validate the effectiveness and versatility of the proposed \n{} framework.


\begin{wraptable}{r}{0.48\textwidth} 
    \setlength{\tabcolsep}{1pt}    
    \vspace{-1\baselineskip}
    \caption{Test mean reciprocal rank (MRR) scores on TGB datasets. TGB leaderboard is \href{https://tgb.complexdatalab.com/docs/leader_linkprop/}{publicly available}.}
    \label{tab:tgb_results}
    \centering
    \footnotesize

    \begin{tabular}{@{}lc|cc@{}}
    \toprule
    \textbf{Datasets} & \textbf{DyGFormer} & \textbf{w/ \n{}} & \textbf{Imp (\%)} \\
    \midrule
    tgbl-wiki & 0.798 (rank 1) ~&~ \textbf{0.815} (rank 1) & {\color{darkgreen}2.13\%} \\
    \midrule
    tgbl-review & 0.224 (rank 6) ~&~ \textbf{0.419} (rank 2) & {\color{darkgreen}87.05\%} \\
    \midrule
    
    tgbl-coin & 0.752 (rank 2) ~&~ \textbf{0.794} (rank 1) & {\color{darkgreen}5.59\%} \\
    \bottomrule
    \end{tabular}    
    \vspace{-1\baselineskip}
\end{wraptable}
\setlength{\tabcolsep}{6pt}


We also present the results of \n{} under the historical (hist) and the inductive (ind) negative sampling strategies in Table~\ref{tab:main_result}. As shown in Table~\ref{tab:main_result}, \n{} consistently improves TGNNs under both negative sampling strategies, achieving improvements of up to 38.48\% and 54.13\% for the historical and inductive negative sampling strategies, respectively. This is because \trc{} maintains the interaction histories of node pairs as the test stage progresses, allowing the underlying models to better capture historical interaction patterns under historical sampling and inductive sampling, more accurately predicting future connections. Please refer to Table~\ref{tab:appendix_full_main_results} for full results.

\noindent\textbf{Performance on three datasets from temporal graph benchmark (TGB).} We further evaluate \n{} on the \textit{tgbl-wiki}, \textit{tgbl-review}, and \textit{tgbl-coin} datasets from TGB~\cite{tgb-dataset}. As shown in Table~\ref{tab:tgb_results}, incorporating DyGFormer into the proposed \n{} framework significantly improves its performance, making it the top-performing model on the \textit{tgbl-wiki} and \textit{tgbl-coin} datasets, and the second-best model on the \textit{tgbl-review} dataset. This is because \lte{} is robust to different levels of skewness in datasets (see Section~\ref{sec:when_log_te_work}) and \trc{} improves prediction accuracy whenever the interaction history between the target node pair are informative in predicting their future link.  Detailed dataset statistics and experimental settings are provided in Section~\ref{appendix_sec_tgb_exp_settings}.

\newcolumntype{X}{>{\fontsize{8}{9}\selectfont}c}
\setlength{\tabcolsep}{2pt}
\begin{table*}[t]
\caption{AP for transductive link prediction.}
\label{tab:ablation}
\centering
\scriptsize

\begin{tabular}{X|XXXXXXXXXXXXXX}
\toprule
\textbf{Methods} & \textbf{CP} & \textbf{CO} & \textbf{EN} & \textbf{FL} & \textbf{LA} & \textbf{MO} & \textbf{RE} & \textbf{SE} & \textbf{UC} & \textbf{UT} & \textbf{UV} & \textbf{US} & \textbf{WK} \\
\midrule
GraphMixer & 75.90 & 91.94 & 82.26 & 90.98 & 75.56 & 82.83 & 97.33 & 93.34 & 93.38 & 62.61 & 52.20 & 71.55 & 97.23 \\

w/ \lte{} & 78.07 & 92.22 & 82.76 & 90.99 & 75.21 & 83.09 & 97.35 & 92.64 & 94.98 & 62.65 & 52.21 & 70.88 & 97.28 \\
{Imp. (\%)} & {\color{darkgreen}2.86\%} & {\color{darkgreen}0.30\%} & {\color{darkgreen}0.61\%} & {\color{darkgreen}0.01\%} & {\color{RawSienna}-0.46\%} & {\color{darkgreen}0.31\%} & {\color{darkgreen}0.02\%} & {\color{RawSienna}-0.75\%} & {\color{darkgreen}1.71\%} & {\color{darkgreen}0.06\%} & {\color{darkgreen}0.02\%} & {\color{RawSienna}-0.94\%} & {\color{darkgreen}0.05\%} \\
\midrule

w/ \trc{} & 75.93 & 95.15 & 89.88 & 96.72 & 88.15 & 83.36 & 98.81 & 93.71 & 94.90 & 62.83 & 57.57 & {71.61} & 98.85 \\
{Imp. (\%)} & {\color{darkgreen}0.04\%} & {\color{darkgreen}3.49\%} & {\color{darkgreen}9.26\%} & {\color{darkgreen}6.31\%} & {\color{darkgreen}16.66\%} & {\color{darkgreen}0.64\%} & {\color{darkgreen}1.52\%} & {\color{darkgreen}0.40\%} & {\color{darkgreen}1.63\%} & {\color{darkgreen}0.35\%} & {\color{darkgreen}10.30\%} & {\color{darkgreen}0.08\%} & {\color{darkgreen}1.67\%} \\
\midrule

w/ \n{} & 78.38 & 95.26 & 90.97 & 96.75 & 88.13 & 83.53 & 98.84 & 93.41 & \underline{96.20} & 62.98 & 57.74 & 71.57 & 98.89 \\
{Imp. (\%)} & {\color{darkgreen}3.27\%} & {\color{darkgreen}3.61\%} & {\color{darkgreen}10.59\%} & {\color{darkgreen}6.34\%} & {\color{darkgreen}16.64\%} & {\color{darkgreen}0.85\%} & {\color{darkgreen}1.55\%} & {\color{darkgreen}0.07\%} & {\color{darkgreen}3.02\%} & {\color{darkgreen}0.59\%} & {\color{darkgreen}10.61\%} & {\color{darkgreen}0.03\%} & {\color{darkgreen}1.71\%} \\

\midrule
\midrule
DyGFormer & 97.91 & 98.31 & 92.46 & 98.92 & 93.01 & 87.66 & 99.22 & 94.66 & 95.66 & 65.07 & 55.88 & 70.44 & 99.02 \\ 
w/ \lte{} & 98.73 & 98.36 & 92.61 & 98.93 & 93.92 & 88.59 & 99.27 & 94.74 & 96.68 & 67.24 & 56.37 & 70.98 & 99.23 \\ 
{Imp. (\%)} & {\color{darkgreen}0.84\%} & {\color{darkgreen}0.05\%} & {\color{darkgreen}0.16\%} & {\color{darkgreen}0.02\%} & {\color{darkgreen}0.98\%} & {\color{darkgreen}1.06\%} & {\color{darkgreen}0.06\%} & {\color{darkgreen}0.08\%} & {\color{darkgreen}1.06\%} & {\color{darkgreen}3.34\%} & {\color{darkgreen}0.88\%} & {\color{darkgreen}0.77\%} & {\color{darkgreen}0.21\%} \\ 

\midrule

w/ \trc{} & 97.57 & 98.65 & 92.59 & 98.92 & 93.44 & 87.55 & 99.23 & 94.72 & 96.00 & 64.53 & 55.99 & 71.23 & 99.07 \\ 
{Imp. (\%)} & {\color{RawSienna}-0.35\%} & {\color{darkgreen}0.34\%} & {\color{darkgreen}0.14\%} & {\color{darkgreen}0.00\%} & {\color{darkgreen}0.47\%} & {\color{RawSienna}-0.13\%} & {\color{darkgreen}0.02\%} & {\color{darkgreen}0.07\%} & {\color{darkgreen}0.36\%} & {\color{RawSienna}-0.83\%} & {\color{darkgreen}0.20\%} & {\color{darkgreen}1.12\%} & {\color{darkgreen}0.05\%} \\ 
\midrule

w/ \n{} & {98.67} & {98.70} & {92.66} & {98.94} & {94.03} & 88.49 & {99.29} & {94.74} & {96.72} & {66.39} & 56.02 & {71.40} & {99.25} \\ 

{Imp. (\%)} & {\color{darkgreen}0.78\%} & {\color{darkgreen}0.40\%} & {\color{darkgreen}0.22\%} & {\color{darkgreen}0.03\%} & {\color{darkgreen}1.10\%} & {\color{darkgreen}0.95\%} & {\color{darkgreen}0.07\%} & {\color{darkgreen}0.08\%} & {\color{darkgreen}1.11\%} & {\color{darkgreen}2.03\%} & {\color{darkgreen}0.25\%} & {\color{darkgreen}1.36\%} & {\color{darkgreen}0.23\%} \\ 

\bottomrule
\end{tabular}
\vskip -0.2in
\end{table*}
\setlength{\tabcolsep}{6pt}

\subsection{Ablation Study}
\label{sec:ablation_study}

We conduct experiments to examine the effectiveness of the proposed \lte{} and \trc{} modules. Table~\ref{tab:ablation} presents the test performance of \n{} and its two variants: \textbf{w/ \lte{}}, where we replace the original TE in TGNNs with the proposed \lte{} and keep the rest unchanged; \textbf{w/ \trc{}}, where we integrate the \trc{} module into TGNNs and keep the rest unchanged. As shown from the results, both \lte{} and \trc{} improve the underlying TGNN performance when integrated individually. The performance further boosts when they are combined together. This indicates that our designed \lte{} and \trc{} are highly effective and versatile across datasets with various domains and temporal scales. We demonstrate in Section~\ref{sec:appendix_inductive_ablation} that \lte{} and \trc{} are also effective in the inductive setting, consistently enhancing the performance of integrated TGNNs.

\subsection{Robustness to the Increase of Negative Links}
\label{sec:exp_more_negative}
\begin{wraptable}{r}{0.55\textwidth} 
\setlength{\tabcolsep}{1pt}
\vspace{-1\baselineskip}
\caption{AP of methods under various numbers of negative links during testing. \textbf{NEG=50} indicates that each positive link is evaluated against 50 negative links in the AP computation.}
\label{tab:exp_more_negative}

\centering
\tiny

\begin{tabular}{@{}c|cccc||c|cccc@{}}
\toprule
\textbf{Method} & \textbf{EN} & \textbf{LA} & \textbf{UC} & \textbf{UV} & \textbf{Method} & \textbf{EN} & \textbf{LA} & \textbf{UC} & \textbf{UV} \\
\hline
\noalign{\vskip 0.015in}
\multicolumn{1}{l}{\textbf{\, NEG = 1}} &  \\ [0.015in]
\hline
GraphMixer  & 82.26 & 75.56 & 93.38 & 52.20 & DyGFormer  & 92.46 & 93.01 & 95.66 & 55.88 \\
w/ \n{}  & 90.97 & 88.13 & 96.20 & 57.74 & {w/ \n{}}  & 92.66 & 94.03 & 96.72 & 56.02 \\ 
Imp (\%)  & {\color{darkgreen}10.59\%} & {\color{darkgreen}16.64\%} & {\color{darkgreen}3.02\%} & {\color{darkgreen}10.61\%} & {Imp (\%)} & {\color{darkgreen}0.22\%} & {\color{darkgreen}1.10\%} & {\color{darkgreen}1.11\%} & {\color{darkgreen}0.25\%} \\
\hline

\noalign{\vskip 0.015in}
\multicolumn{1}{l}{\textbf{\, NEG = 5}} &  \\ [0.015in]
\hline
GraphMixer  & 53.42 & 47.06 & 82.05 & 18.17 & DyGFormer  & 75.81 & 77.77 & 88.77 & 20.86 \\
w/ \n{}  & 70.36 & 65.05 & 88.40 & 21.62 & w/ \n{}  & 76.37 & 80.58 & 90.76 & 20.97 \\
Imp (\%)  & {\color{darkgreen}31.72\%} & {\color{darkgreen}38.24\%} & {\color{darkgreen}7.73\%} & {\color{darkgreen}18.99\%} & Imp (\%)  & {\color{darkgreen}0.74\%} & {\color{darkgreen}3.61\%} & {\color{darkgreen}2.24\%} & {\color{darkgreen}0.55\%} \\
\hline
\noalign{\vskip 0.015in}
\multicolumn{1}{l}{\textbf{\, NEG = 25}} &  \\ [0.015in]
\hline
GraphMixer & 22.77 & 23.77 & 63.54 & 4.36 & DyGFormer  & 45.15 & 51.58 & 78.96 & 5.13 \\
w/ \n{} & 36.57 & 34.97 & 73.26 & 5.32 & w/ \n{}  & 46.39 & 56.61 & 81.74 & 5.28 \\
Imp (\%)  & {\color{darkgreen}60.61\%} & {\color{darkgreen}47.12\%} & {\color{darkgreen}15.30\%} & {\color{darkgreen}22.04\%} & Imp (\%)  & {\color{darkgreen}2.74\%} & {\color{darkgreen}9.74\%} & {\color{darkgreen}3.52\%} & {\color{darkgreen}2.92\%} \\
\hline
\noalign{\vskip 0.015in}
\multicolumn{1}{l}{\textbf{\, NEG = 50}} &  \\ [0.015in]
\hline
GraphMixer  & 14.09 & 16.80 & 54.14 & 2.25 & DyGFormer & 31.06 & 35.64 & 73.47 & 2.65 \\
w/ \n{}  & 23.43 & 24.66 & 63.74 & 2.74 & w/ \n{}  & 32.43 & 44.76 & 76.59 & 2.72 \\
Imp (\%)  & {\color{darkgreen}66.29\%} & {\color{darkgreen}46.76\%} & {\color{darkgreen}17.74\%} & {\color{darkgreen}21.78\%} & Imp (\%)  & {\color{darkgreen}4.41\%} & {\color{darkgreen}25.59\%} & {\color{darkgreen}4.25\%} & {\color{darkgreen}2.64\%} \\

\bottomrule
\end{tabular}

\vspace{-1\baselineskip}
\end{wraptable}
\setlength{\tabcolsep}{6pt}
In this experiment, we evaluate the robustness of \n{} against an increasing number of negative links per positive link. A negative link refers to a pair of nodes that are not currently connected and are used as negative samples in the link prediction process. Ideally, a link prediction model should assign a higher connection probability to positive links and a probability close to zero to negative links. \emph{The more negative links, the more difficult the task is.} In the default setting, the number of negative links is set to 1 per positive link. We run the experiments on four representative datasets (EN, LA, UC, and UV), and report the results in Table~\ref{tab:exp_more_negative}. ``NEG=50'' denotes that each positive link is evaluated against 50 negative links when computing its connection probability. 

As shown in Table~\ref{tab:exp_more_negative}, integrating GraphMixer and DyGFormer into \n{} consistently enhances their performance across different numbers of negative links per positive link. Furthermore, the improvement ratio steadily increases as the number of negative links grows. For example, on the EN dataset, the improvement ratio for GraphMixer increases from 10.59\% to 66.29\% as the number of negative links per positive link increases from 1 to 50. This is because leveraging the interaction histories stored in \trc{} between target node pairs helps to avoid predicting negative links as positive ones. These results suggest that the proposed \n{} framework effectively improves TGNN performance in scenarios where the ratio of negative links to positive links is high, reflecting conditions that are more practical for sparse CTTGs.

\subsection{Adaptivity to Different TGNN Architectures}
\label{sec:portability_main_text}
\begin{wraptable}{r}{0.55\textwidth} 
\vspace{-1\baselineskip}
\setlength{\tabcolsep}{1pt}
\caption{AP Improvement (Imp.\%) of various TGNNs when integrated into \n{}. The format is of AP~({\color{blue} +Imp.\%}). Please refer to Table~\ref{tab:portability_appendix} for full results.}
\label{tab:portability_main_text}

\centering
\tiny

\begin{tabular}{@{}c|cccc@{}}
\toprule
\textbf{Method (w/ \n{})} &  \textbf{EN} & \textbf{LA} & \textbf{UC} & \textbf{UV} \\
\midrule
{CAWN}  & 91.23 ({\color{darkgreen}+3.21\%}) & 91.02 ({\color{darkgreen}+4.64\%}) & 96.69 ({+\color{darkgreen}1.81\%}) & 57.49 ({\color{darkgreen}+8.72\%}) \\

{TGAT}  & 91.37 ({\color{darkgreen}+25.35\%}) & 91.60 ({\color{darkgreen}+24.85\%}) & 96.36 ({\color{darkgreen}+21.53\%}) & 60.03 ({\color{darkgreen}+12.98\%})\\

{TGN}  & 92.34 ({\color{darkgreen}+6.03\%}) & 92.83 ({\color{darkgreen}+22.58\%}) & 95.36 ({\color{darkgreen}+3.80\%}) & 67.80 ({\color{darkgreen}+3.22\%}) \\

{JODIE}  & 90.62 ({\color{darkgreen}+6.88\%} ) & 87.95 ({\color{darkgreen}+25.33\%}) & 92.44 ({\color{darkgreen}+3.68\%}) & 65.57 ({\color{darkgreen}+3.22\%})\\
\bottomrule
\end{tabular}
\vspace{-1\baselineskip}
\end{wraptable}
\setlength{\tabcolsep}{6pt}
We below show the adaptivity of \n{} to different TGNN architectures. Specifically, we integrate a random walk-based TGNN, i.e. CAWN~\cite{CAWN}, and three temporal neighbor-based TGNNs, i.e., TGAT~\cite{tgat}, TGN~\cite{TGN}, and JODIE~\cite{JODIE}, into \n{}. Since JODIE does not utilize the time encoding function, we report its performance with \trc{} integrated. As shown in Table~\ref{tab:portability_main_text}, \n{} consistently improves their performance, indicating the effectiveness of \n{} for different TGNN architectures. Additionally, we show in Table~\ref{tab:portability_appendix} that applying \lte{} and \trc{} individually also boosts the performance of the underlying TGNNs.

\begin{figure*}[t!]
    \centering
    \begin{subfigure}[t]{0.49\textwidth}
        \centering
        \includegraphics[width=\textwidth]{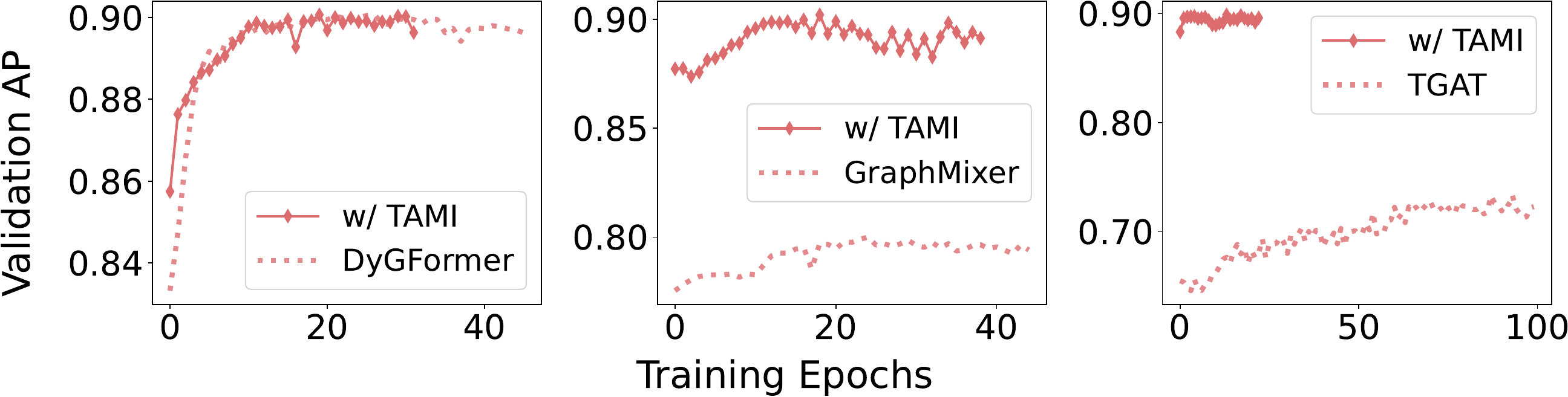}
        \caption{The EN dataset}
        \label{fig:converge_speed_AP_enron}
    \end{subfigure}%
    \hspace{0.012\textwidth}
    \begin{subfigure}[t]{0.49\textwidth}
        \centering
        \includegraphics[width=\textwidth]{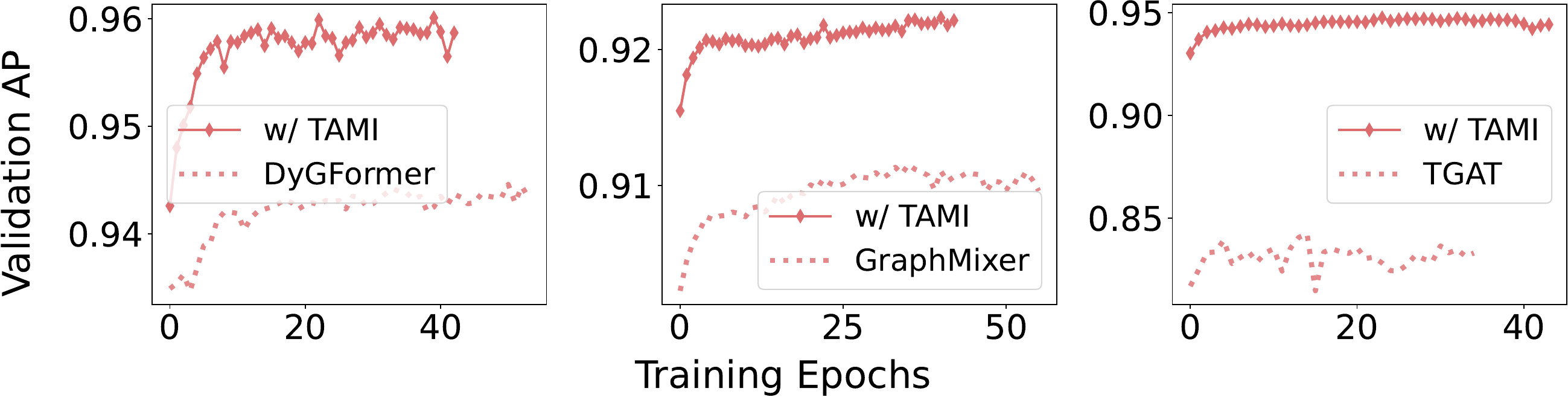}
        \caption{The UC dataset}
        \label{fig:converge_speed_AP_uci}
    \end{subfigure}
    \caption{Validation AP vs. training epochs on the (a) EN and (b) UC datasets. \n{} enables TGNNs to achieve higher validation average precision with fewer training epochs.}
\label{fig:converge_speed}
\vskip -0.2in
\end{figure*}

\subsection{Improved Training Efficiency}
\label{sec:exp_trainnig_acceleration_exp}

\begin{wrapfigure}{r}{.65\textwidth} 
    \begin{center} 
    \vspace{-1\baselineskip}
    \centerline{\includegraphics[width=0.65\columnwidth]{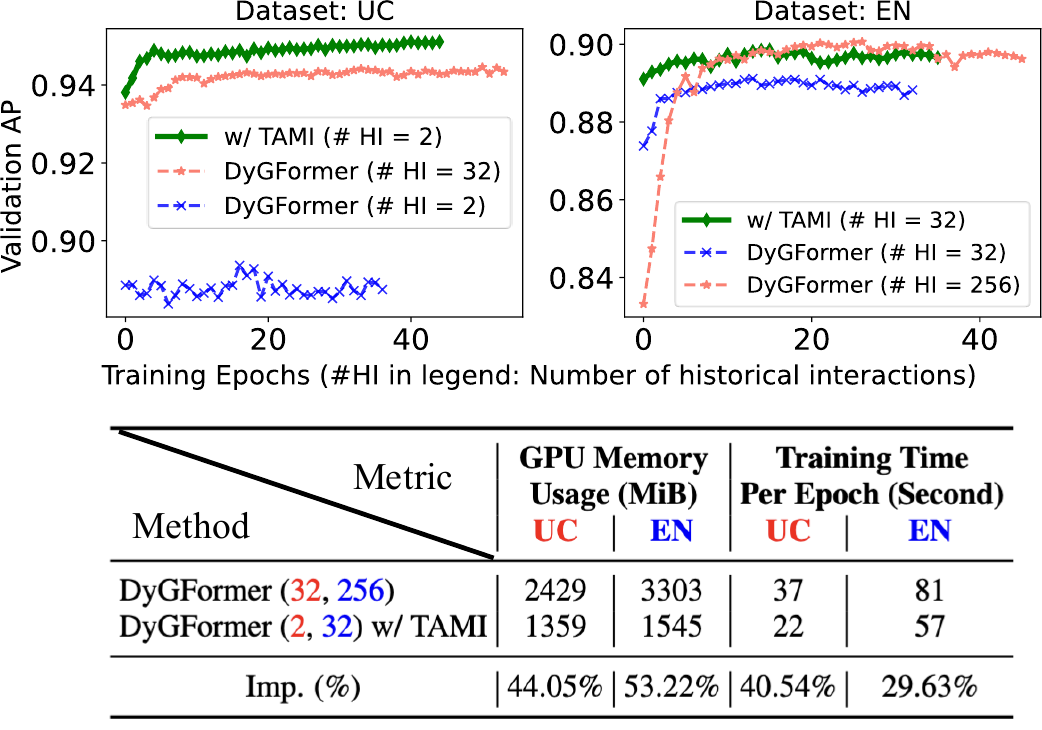}}
    \caption{DyGFormer integrated with \n{} can compute temporal node embeddings using significantly fewer historical interactions while achieving better performance.}    
    \label{fig:converge_dyg_different_hist}
    \end{center}    
    \vspace{-2\baselineskip}
\end{wrapfigure}

We observe that \n{} can speed up the training of underlying TGNNs. To show this, we plot the validation AP of TGNNs during training in Figure~\ref{fig:converge_speed}. On the EN and UC datasets, TGNNs within the \n{} framework achieve higher validation AP scores with \emph{much fewer} training epochs, compared to their vanilla counterparts. In other words, the underlying models converge faster than their vanilla versions. These findings suggest that \n{} can speed up the convergence of TGNNs while yielding improved validation AP scores. This is because \lte{} facilitates the learning of frequency parameters in the time encoding function by balancing input temporal differences, while \trc{} enables the underlying model to more effectively capture the patterns of historical interactions between target node pairs.

In addition, we demonstrate that DyGFormer, when integrated into our \n{} framework, can compute temporal node embeddings using \emph{significantly fewer} historical interactions while achieving equal or even better performance. As shown in Figure~\ref{fig:converge_dyg_different_hist}, when integrated into \n{}, DyGFormer on the UC and EN datasets requires only 2 and 32 historical interactions for attention calculation, respectively, outperforming the vanilla counterpart that uses 16$\times$ and 8$\times$ more historical interactions. The reduced number of historical interactions results in lower GPU memory usage (up to 53.22\%). Moreover, on both datasets, DyGFormer under the \n{} framework converges much faster than the vanilla version, with fewer training epochs and up to 40.54\% reduction in training time per epoch.

\section{Conclusion}
\label{sec/concllusion}
In this paper, we observed the presence of heterogeneity in temporal interactions of CTTGs and proposed a novel framework \n{} to address the challenges therein. \n{} has two main modules, namely \lte{} and \trc{}. \lte{} {balances the skewness in the time differences via a simple yet effective logarithmic transformation}, and \trc{} prevents the historical interactions for each target node pair from being forgotten. Existing temporal graph neural networks can be seamlessly and readily integrated into \n{}. Extensive experiments on 13 classic open datasets and three TGB datasets show that \n{} substantially improves the link prediction accuracy of the underlying models as well as their training efficiency, in both transductive and inductive settings. 

\begin{ack}
We thank the anonymous reviewers for their constructive feedback. This work was supported in part by the Guangdong Provincial Key Laboratory of IRADS (2022B1212010006), the Guangdong Higher Education Upgrading Plan (2021-2025), and the Guangdong and Hong Kong Universities ``1+1+1'' Joint Research Collaboration Scheme. This work was also supported in part by the National Science Foundation under Grant No. IIS-2209921, and the International Energy Joint R\&D Program of the Korea Institute of Energy Technology Evaluation and Planning (KETEP), granted financial resource from the Ministry of Trade, Industry \& Energy, Republic of Korea (No. 20228530050030).
\end{ack}

\bibliography{main}
\bibliographystyle{plain}


\clearpage
\section*{NeurIPS Paper Checklist}

\begin{enumerate}

\item {\bf Claims}
    \item[] Question: Do the main claims made in the abstract and introduction accurately reflect the paper's contributions and scope?
    \item[] Answer: \answerYes{} 
    \item[] Justification: This is the first study to identify the presence of heterogeneity in temporal interactions of CTTGs and investigate its impact on link prediction performance. We propose a novel framework \n{} to handle the heterogeneity in temporal interactions, and existing graph neural networks can be seamlessly integrated into our \n{} framework. The claims in the paper are supported by related work discussion in Section~\ref{sec:related_work}, experiments in Section~\ref{sec/experiments}, and mathematical derivations in Section~\ref{sec:proof_prop_1}. Please also refer to the additional experiments in the Appendix.

\item {\bf Limitations}
    \item[] Question: Does the paper discuss the limitations of the work performed by the authors?
    \item[] Answer: \answerYes{} 
    \item[] Justification: Please refer to Section~\ref{sec:when_log_te_work}.

\item {\bf Theory assumptions and proofs}
    \item[] Question: For each theoretical result, does the paper provide the full set of assumptions and a complete (and correct) proof?
    \item[] Answer: \answerYes{} 
    \item[] Justification: We provide the mathematical steps and justification of Proposition 1 in Section~\ref{sec:proof_prop_1}.

    \item {\bf Experimental result reproducibility}
    \item[] Question: Does the paper fully disclose all the information needed to reproduce the main experimental results of the paper to the extent that it affects the main claims and/or conclusions of the paper (regardless of whether the code and data are provided or not)?
    \item[] Answer: \answerYes{} 
    \item[] Justification: We provide a detailed description of our experimental setup in Section~\ref{appendix:implementation_details}, Section~\ref{appendix:config_baselines}, and Section~\ref{appendix_sec_tgb_exp_settings}. We also include our code in the supplementary material.

\item {\bf Open access to data and code}
    \item[] Question: Does the paper provide open access to the data and code, with sufficient instructions to faithfully reproduce the main experimental results, as described in supplemental material?
    \item[] Answer: \answerYes{} 
    \item[] Justification: Dataset sources are provided in Section~\ref{appendix:dataset_description} and Section~\ref{appendix_sec_tgb_exp_settings}. Our code is provided in the supplemental material.

\item {\bf Experimental setting/details}
    \item[] Question: Does the paper specify all the training and test details (e.g., data splits, hyperparameters, how they were chosen, type of optimizer, etc.) necessary to understand the results?
    \item[] Answer: \answerYes{} 
    \item[] Justification: We describe dataset splits in Section~\ref{sec:exp_settings}, hyperparameter settings in Section~\ref{appendix:config_baselines}, and detailed implementations in Section~\ref{appendix:implementation_details}.

\item {\bf Experiment statistical significance}
    \item[] Question: Does the paper report error bars suitably and correctly defined or other appropriate information about the statistical significance of the experiments?
    \item[] Answer: \answerYes{}
    \item[] Justification: We report the mean and standard deviation of results on all our
experiments, over 5 repeated runs. We state these details explicitly in the result tables. Due to the limited space, standard deviations of the results in the main text tables are provided in Section~\ref{sec:std_main_text}.

\item {\bf Experiments compute resources}
    \item[] Question: For each experiment, does the paper provide sufficient information on the computer resources (type of compute workers, memory, time of execution) needed to reproduce the experiments?
    \item[] Answer: \answerYes{} 
    \item[] Justification: Computational resources are described in Section~\ref{appendix:implementation_details}. We provided a detailed space complexity discussion at the end of Section~\ref{sec:model_trc}.
    
\item {\bf Code of ethics}
    \item[] Question: Does the research conducted in the paper conform, in every respect, with the NeurIPS Code of Ethics \url{https://neurips.cc/public/EthicsGuidelines}?
    \item[] Answer: \answerYes{} 
    \item[] Justification:  Our proposed framework is widely applicable to science and engineering applications and does not have any conflict with the NeurIPS code of Ethics.

\item {\bf Broader impacts}
    \item[] Question: Does the paper discuss both potential positive societal impacts and negative societal impacts of the work performed?
    \item[] Answer: \answerNA{} 
    \item[] Justification: The proposed method can be applied to a wide variety of science and engineering applications. There is no clear path to a negative societal impact from our work.
Guidelines:
    
\item {\bf Safeguards}
    \item[] Question: Does the paper describe safeguards that have been put in place for responsible release of data or models that have a high risk for misuse (e.g., pretrained language models, image generators, or scraped datasets)?
    \item[] Answer: \answerNA{} 
    \item[] Justification: Our paper does not include data or models with a high risk for misuse.

\item {\bf Licenses for existing assets}
    \item[] Question: Are the creators or original owners of assets (e.g., code, data, models), used in the paper, properly credited and are the license and terms of use explicitly mentioned and properly respected?
    \item[] Answer: \answerYes{} 
    \item[] Justification: All benchmarks used in the paper are properly cited.

\item {\bf New assets}
    \item[] Question: Are new assets introduced in the paper well documented and is the documentation provided alongside the assets?
    \item[] Answer: \answerYes{} 
    \item[] Justification: Code has been submitted with the paper and includes clear documentation
for running the experiments.

\item {\bf Crowdsourcing and research with human subjects}
    \item[] Question: For crowdsourcing experiments and research with human subjects, does the paper include the full text of instructions given to participants and screenshots, if applicable, as well as details about compensation (if any)? 
    \item[] Answer: \answerNA{} 
    \item[] Justification: This paper does not involve crowdsourcing nor research with human subjects.

\item {\bf Institutional review board (IRB) approvals or equivalent for research with human subjects}
    \item[] Question: Does the paper describe potential risks incurred by study participants, whether such risks were disclosed to the subjects, and whether Institutional Review Board (IRB) approvals (or an equivalent approval/review based on the requirements of your country or institution) were obtained?
    \item[] Answer: \answerNA{} 
    \item[] Justification: This paper does not involve crowdsourcing nor research with human subjects.

\item {\bf Declaration of LLM usage}
    \item[] Question: Does the paper describe the usage of LLMs if it is an important, original, or non-standard component of the core methods in this research? Note that if the LLM is used only for writing, editing, or formatting purposes and does not impact the core methodology, scientific rigorousness, or originality of the research, declaration is not required.
    \item[] Answer: \answerNA{} 
    \item[] Justification: The core method development in this research does not involve LLMs as any important, original, or non-standard components.

\end{enumerate}

\clearpage
\appendix
\section*{Technical Appendix and Supplementary Material}

\setlength{\tabcolsep}{4pt}

In the Appendix, we provide additional supplementary material to the main paper. The structure is
as follows:

\begin{enumerate}
    \item Section~\ref{sec:proof_prop_1} outlines the proof of Proposition 1.
    \item Section~\ref{sec:appendix} outlines the experimental settings in detail.
    \item Section~\ref{sec:appendix_transductive_results} presents additional results for link prediction in the transductive setting.
    \item Section~\ref{appendix:inductive_link_prediction} shows the results in the inductive setting.
    \item Section~\ref{appendix_sec_tgb_exp_settings} outlines the experimental settings on TGB datasets.
    \item Section~\ref{sec:std_main_text} presents the standard deviations of the results in the main text tables.
\end{enumerate}

\section{Proof of Proposition 1}
\label{sec:proof_prop_1}

Suppose $\Delta t$ follows a Pareto distribution with the shape parameter $\alpha > 3$, whose skewness is always greater than 2. Note that if the shape parameter is $0 < \alpha \leq 2$, the variance is infinite, so the skewness is undefined. Also, if the shape parameter is $2 < \alpha \leq 3$, its third moment is infinite. Thus, we focus on the case of $\alpha > 3$. Then, for any value of $\alpha > 3$, we show that \lte{} always reduces the skewness of $\Delta t$ to 2.

\begin{proof}
Let $T$ be a random variable that represents the temporal difference $\Delta t$ and follows a Pareto distribution. Then, $T + 1$ also follows a (shifted) Pareto distribution. Letting $X = T + 1$, we can write the probability density function (PDF) of $X$ as
$$
f_X(x) = \frac{\alpha x_{\text{min}}^\alpha}{x^{\alpha + 1}}, \quad x \geq x_{\text{min}},
$$
where $\alpha > 3$ is the shape parameter, and $x_{\text{min}} > 1$ is the scale parameter.

First, we show that the skewness $\Gamma$ of $X$ is always greater than 2. We write the skewness $\Gamma$ of $X$ as a function of $\alpha$, i.e., 
\begin{equation}
\label{proof_1}
\Gamma = g(\alpha) = \frac{2(1+\alpha)}{\alpha - 3} \sqrt{1 - \frac{2}{\alpha}}
\end{equation}
It is then straightforward to see that $g(\alpha)$ is monotonically decreasing, with $g(\alpha) \rightarrow 2$ as $\alpha \rightarrow \infty$. To show that $g(\alpha)$ is monotonically decreasing, we can show that $\ln g(\alpha)$ exists and $\ln g(\alpha)$ is monotonically decreasing. We first take the natural logarithm on both sides of (\ref{proof_1}) and differentiate $\ln g(\alpha)$ with respect to $\alpha$. We can then show that $\frac{g'(\alpha)}{g(\alpha)} < 0$. In addition, we can easily see from (\ref{proof_1}) that $g(\alpha) \rightarrow 2$ as $\alpha \rightarrow \infty$. Therefore, the skewness $\Gamma$ of $X$ is always greater than 2.

Next, for any value of $\alpha > 3$, we show that \lte{} always reduces the skewness $\Gamma$ to 2. Recall that \lte{} transforms $X$ via a logarithmic function $Y \!=\! \ln(X)$. By using the change-of-variables formula for probability distributions, we can obtain the PDF of $Y$ as follows:
\begin{equation*}
    f_{Y}(y) = f_X(x) \cdot \left |\frac{dx}{dy} \right | = \alpha x_{\text{min}}^\alpha e^{-\alpha y}.
\end{equation*}
Letting $\mu = \ln(x_{\text{min}}) > 0$, we have 
\begin{equation*}
f_Y(y) = \alpha e^{-\alpha (y - \mu)}, \quad y \geq \mu,
\end{equation*}
which is a shifted exponential distribution with the rate parameter $\alpha > 3$. Since the skewness of a (shifted) exponential distribution is 2 regardless of the value of the rate parameter $\alpha$, \lte{} always reduces the skewness to 2.  
\end{proof}

\section{Experimental Settings}
\label{sec:appendix}

\subsection{Descriptions of Datasets}
\label{appendix:dataset_description}
We use 13 datasets collected by~\cite{edgebank} in our experiments.

\begin{enumerate}
    \item \textbf{Can. Parl. (CP)} is a dynamic political network that captures the interactions among Canadian Members of Parliament (MPs) from 2006 to 2019. Each node represents an MP from an electoral district, and an edge is established when two MPs cast a ``yes'' vote on the same bill. The weight of each edge reflects the annual frequency with which one MP votes ``yes'' alongside another.
    \item \textbf{Contact (CO)} describes how the physical proximity evolves among about 700 university students over a month. Each student has a unique ID and edges between students denote that they are within close proximity to each other. Each edge is assigned a weight that reflects the physical proximity between students.
    \item \textbf{Enron (EN)} is an email correspondence dataset that records the emails exchanged among employees of the ENRON energy company over three years.
    \item \textbf{Flights (FL)} is a dynamic flight network illustrating the development of air traffic during the COVID-19 pandemic. Nodes represent airports and the tracked flights are denoted as edges. The edge weights reflect the number of flights between two airports in a day.
    \item \textbf{LastFM (LA)} records the interaction between users and songs. Users and songs are nodes and edges between them represent a user-listens-to-song relation. The dataset contains no attributes.
    \item \textbf{Mooc (MO)} is a dataset that captures students' interactions with online course materials. Each edge represents a student accessing a content unit and is associated with a 4-dimensional feature vector.
    \item \textbf{Reddit (RE)} comprises user posts submitted to subreddits over one month. Users and subreddits are nodes, while timestamped posting requests form the edges. Edge features are LIWC-feature vectors~\cite{LIWC} of edit texts.
    \item \textbf{Social Evo. (SE)} is a mobile phone proximity network that tracks the daily interactions of an undergraduate dormitory over eight months. Each edge is associated with a 2-dimensional feature vector.
    \item \textbf{UCI (UC)} is an unattributed online communication network among university students. Nodes are university students and edges are messages posted by students.
    \item \textbf{UN Trade (UT)} is a food and agriculture trading graph between 181 nations for more than 30 years. The edge weights indicate the total sum of normalized agriculture import or export values between two countries.
    \item \textbf{UN Vote (UV)} captures roll-call voting behavior in the United Nations General Assembly. Each time two nations vote a ``yes'' on the same item, the edge weight between them is incremented by one.
    \item \textbf{US Legis. (US)} is a senate co-sponsorship graph that captures the social dynamics among US legislators. The edge weights specify the number of times two congresspersons have co-sponsored a bill in a given Congress.
    \item \textbf{Wikipedia (WK)} records the edits on Wikipedia pages over a month. Editors and Wiki pages are modeled as nodes, and posting requests are timestamped edges. Edge features are 172-dimensional LIWC feature vectors~\cite{LIWC}.
\end{enumerate}

We present the dataset statistics in Table~\ref{tab:appendix_dataset_stat}, where ``\#N\&E Feat'' refers to the dimensions of node and raw edge features. {Table~\ref{tab:appendix_skew_datasets} summarizes the skewness of all 13 datasets, where the skewness is measured for the interaction intervals between pairs of nodes in each dataset.}

\begin{table*}[t]
\caption{Statistics of Datasets. The `-' symbol denotes that the dataset does not contain the corresponding feature.}
\label{tab:appendix_dataset_stat}
\centering
\scriptsize

\begin{tabular}{c|cccccccc}
\toprule

\textbf{Datasets} & Domains & \#Nodes & \#Edges & \# Unique Edges & \#N\&E Feat & Duration & Unique Steps & Time Granularity \\
\midrule

\textbf{CP}    & Politics & 734 &  74,478 & 51,331 & -- \& 1 & 14 years &  14 & years \\
\textbf{CO}  & Proximity &  694 & 2,426,280 & 79,531 &  -- \& 1 & 1 month & 8,065 &  5 minutes\\
\textbf{EN} & Social & 184 &  125,235 & 3,125 & -- \& -- & 3 years & 22,632 & Unix timestamps \\
\textbf{FL} & Transport & 13,169 & 1,927,145 & 395,072 &  -- \& 1 & 4 months & 122 & days \\
\textbf{LA} & Interaction & 1,980 &  1,293,103 & 154,993 & -- \& -- & 1 month & 1,283,614 & Unix timestamps \\
\textbf{MO} & Interaction & 7,144 & 411,749 &  178,443 & -- \& 4 & 17 months & 345,600 & Unix timestamps \\
\textbf{RE} & Social & 10,984 &  672,447 &  78,516 &  -- \& 172 & 1 month &  669,065 & Unix timestamps \\
\textbf{SE} & Proximity & 74 & 2,099,519 & 4,486 & -- \& 2 & 8 months & 565,932 & Unix timestamps \\
\textbf{UC} & Social & 1,899 & 59,835 & 20,296 & -- \& -- & 196 days & 58,911 & Unix timestamps \\
\textbf{UT} & Economics &  255 & 507,497 & 36,182 & -- \& 1 & 32 years & 32 & years \\
\textbf{UV} & Politics & 201 & 1,035,742 &  31,516 & -- \& 1 & 72 years & 72 & years \\
\textbf{US} & Politics & 225 & 60,396 & 26,423 & -- \& 1 & 12 congresses & 12 & congresses \\
\textbf{WK} & Social & 9,227 & 157,474 & 18,257 & -- \& 172 & 1 month & 152,757 & Unix timestamps \\

\bottomrule
\end{tabular}

\end{table*}

\begin{table*}[t]
\centering
\scriptsize
\caption{Skewness of datasets measured by interaction intervals.}
\begin{tabular}{c|ccccccccccccc} 
\toprule
Datasets & CP	& CO &	EN & FL	& LA & MO & RE & SE &  UC & UT & UV & US & WK \\
\midrule
Skewness & 1.71	& 10.96 & 	4.24	& 2.37	& 3.15	& 4.33 & 2.46  & 18.63	& 5.2 &	6.35 & 	6.36 &	9.17 &	4.48\\
\bottomrule
\end{tabular}
\vskip -0.15in
\label{tab:appendix_skew_datasets}
\end{table*}


\subsection{Descriptions of Baselines}
\label{appendix:baseline_descriptions}
We select nine baseline link prediction methods, covering a wide range of underlying TGNN architectures: random walk-based TGNN (CAWN), and temporal neighbor-based TGNNs (TGN, EdgeBank, JODIE, DyRep, TGAT, TCL, GraphMixer, DyGFormer).

\begin{enumerate}
    \item \textbf{TGN} maintains an evolving memory for each node in a temporal graph, updating its memory when the node participates in an interaction. The stored historical states of the node are subsequently used by an embedding module to compute its future representation.
    \item \textbf{EdgeBank} EdgeBank is a transductive link prediction method with no trainable parameters. It stores observed interactions between nodes in a memory unit, which is updated using various strategies. A future interaction is predicted as positive if it is retained in memory, and negative otherwise \cite{edgebank}. Depending on the memory update strategies, EdgeBank has four variants: EdgeBank$_\infty$ uses unlimited memory and retains all observed edges; EdgeBank$_{tw-ts}$ and EdgeBank$_{tw-re}$ retain only recent edges within a fixed-size time window. The window size for EdgeBank$_{tw-ts}$ is set to the duration of the test split, while EdgeBank$_{tw-re}$ adjusts the window size based on the time intervals between repeated edges; EdgeBank$_{th}$ retains only edges that appear more than a specified threshold number of times. We evaluate all four variants and report the best-performing one.
    \item \textbf{JODIE} is designed for temporal bipartite networks involving user-item interactions. It maintains the states of both user and item nodes and utilizes two coupled recurrent neural networks to update these node states. Additionally, a projection operation is introduced to learn the future representation trajectory for each user and item \cite{JODIE}.
    \item \textbf{DyRep} introduces a recurrent architecture to update node states at each interaction, complemented by a temporal-attentive aggregation module that captures the evolving structural information in temporal graphs. \cite{DyRep}.
    \item \textbf{CAWN} first extracts multiple causal anonymous walks for each node, enabling the exploration of the causality in network dynamics. Then, it employs recurrent neural networks to encode each walk and aggregates them to obtain the final node representation \cite{CAWN}.
    \item \textbf{TGAT} It computes the representation of a node by aggregating its temporal neighbors using the graph attention mechanism~\cite{gat}, with a time encoding function to capture temporal patterns~\cite{tgat}.
    \item \textbf{TCL} first performs a breadth-first search on the temporal subgraph to identify the temporal neighbors of the target node. Subsequently, a graph transformer is used to encode neighbor embeddings and graph topologies, enabling the computation of node representations. Furthermore, a cross-attention mechanism is employed to capture the interdependencies between the two interacting nodes \cite{TCL}.
    \item \textbf{GraphMixer} utilizes the MLP-Mixer \cite{MLPMixer} to encode both temporal information and the historical interactions of the target node. Additionally, a node encoder is employed to aggregate the node features of temporal neighbors \cite{graphmixer}. In their experiments, they show that a fixed time encoding function outperforms its trainable counterpart.
    \item \textbf{DyGFormer} is a Transformer-based architecture that computes node representations by aggregating features from each node’s temporal neighbors. It introduces a neighbor co-occurrence encoding scheme to capture the correlations between nodes within an interaction and a patching technique to help the model capture long-term dependencies \cite{DyGFormer}.
\end{enumerate}


\subsection{Implementation Details}
\label{appendix:implementation_details}
We train all TGNNs (excluding EdgeBank, which has no trainable parameters) using the Adam optimizer~\cite{adam}, with binary cross-entropy loss as the objective function. TGNNs are trained for 100 epochs with early stopping, where the patience score is set to 20. We use a learning rate of 0.0001 and a batch size of 200 for all methods and datasets. The model with the best validation performance is selected for testing. Each method is run \emph{five} times using random seeds ranging from 0 to 4, with the average performance reported to minimize any deviations. For the hyperparameter $\gamma$ in our \trc{} module, we search for the best $\gamma$ range from $0.0001$ to $1$ during the training and validation phases and then use the $\gamma$ with the best validation performance in the test phase. Specifically, we set $\gamma = 0.0001$ for the \textbf{MO} and \textbf{SE} datasets, $\gamma = 0.1$ for the \textbf{UV} dataset, and $\gamma = 0.9$ for the remaining ten datasets. The dimension of historical edge embedding is set equal to the dimension of temporal node embedding, i.e., $d_r = d$, where detailed configurations can be found in Section~\ref{appendix:config_baselines}. 

All our experiments are conducted on a GPU server running Ubuntu 22.04, with PyTorch 2.1.0
and CUDA 12.1. We train and test the proposed \n{} framework using a single NVIDIA A100 80G GPU.

\subsection{Configurations of Baselines}
\label{appendix:config_baselines}
\cite{DyGFormer} conducted an extensive hyperparameter search across all 13 datasets. For consistency, we adopt the optimal hyperparameter settings reported in \cite{DyGFormer} for all baseline methods. We first outline the configurations that remain consistent across all datasets, followed by the specific hyperparameter settings for each dataset.

The consistent configurations are as follows:
\begin{itemize}
    \item \textbf{JODIE} 
        \begin{enumerate}
            \item Dimension of node memory: 172
            \item Dimension of output representation: 172
            \item Memory updater: vanilla recurrent neural network
        \end{enumerate}
    \item \textbf{DyRep}
        \begin{enumerate}
            \item Dimension of time encoding: 100
            \item Dimension of node memory: 172
            \item Dimension of output representation: 172
            \item Number of graph attention heads: 2
            \item Number of graph convolution layers: 1
            \item Memory updater: vanilla recurrent neural network
        \end{enumerate}
    \item \textbf{TGAT}
        \begin{enumerate}
            \item Dimension of time encoding: 100
            \item Dimension of output representation: 172
            \item Number of graph attention heads: 2
            \item Number of graph convolution layers: 2
        \end{enumerate}
    \item \textbf{TGN}
        \begin{enumerate}
            \item Dimension of time encoding: 100
            \item Dimension of node memory: 172
            \item Dimension of output representation: 172
            \item Number of graph attention heads: 2
            \item Number of graph convolution layers: 1
            \item Memory updater: gated recurrent unit~\cite{gru}
        \end{enumerate}
    \item \textbf{CAWN}
        \begin{enumerate}
            \item Dimension of time encoding: 100
            \item Dimension of position encoding: 172
            \item Dimension of output representation: 172
            \item Number of attention heads for encoding walks: 8
            \item Length of each walk (including the target node): 2
            \item Time scaling factor $\alpha$: 1e-6
        \end{enumerate}
    \item \textbf{TCL}
        \begin{enumerate}
            \item Dimension of time encoding: 100
            \item Dimension of depth encoding: 172
            \item Dimension of output representation: 172
            \item Number of attention heads: 2
            \item Number of Transformer layers: 2
        \end{enumerate}
    \item \textbf{GraphMixer}
        \begin{enumerate}
            \item Dimension of time encoding: 100
            \item Dimension of output representation: 172
            \item Number of MLP-Mixer layers: 2
            \item Time gap $T$: 2000
        \end{enumerate}
    \item \textbf{DyGFormer}
        \begin{enumerate}
            \item Dimension of time encoding: 100
            \item Dimension of neighbor co-occurrence encoding $d_C$: 50
            \item Dimension of aligned encoding $d$: 50
            \item Dimension of output representation: 172
            \item Number of attention heads: 2   
            \item Number of Transformer layers: 2
        \end{enumerate}
\end{itemize}

The hyperparameter settings for each method across different datasets are shown in Table~\ref{tab:hyper_config_dropout_rate}, Table~\ref{tab:hyper_config_temporal_neighbor_sampling_size}, and Table~\ref{tab:hyper_config_smaple_strategy}.

\begin{table*}[h!]
\caption{Dropout rates of different methods.}
\label{tab:hyper_config_dropout_rate}
\centering
\scriptsize
\begin{tabular}{c|cccccccc}
\toprule
\textbf{Datasets} & \textbf{JODIE} & \textbf{DyRep} & \textbf{TGAT} & \textbf{TGN} & \textbf{CAWN} & \textbf{TCL} & \textbf{GraphMixer} & \textbf{DyGFormer} \\
\midrule
{CP} & 0.0 & 0.0 & 0.2 & 0.3 & 0.0 & 0.2 & 0.2 & 0.1 \\
{CO} & 0.1 & 0.0 & 0.1 & 0.1 & 0.1 & 0.0 & 0.1 & 0.0 \\
{EN} & 0.1 & 0.0 & 0.2 & 0.0 & 0.1 & 0.1 & 0.5 & 0.0 \\
{FL} & 0.1 & 0.1 & 0.1 & 0.1 & 0.1 & 0.1 & 0.2 & 0.1 \\
{LA} & 0.3 & 0.0 & 0.1 & 0.3 & 0.1 & 0.1 & 0.0 & 0.1 \\
{MO} & 0.2 & 0.0 & 0.1 & 0.2 & 0.1 & 0.1 & 0.4 & 0.1 \\
{RE} & 0.1 & 0.1 & 0.1 & 0.1 & 0.1 & 0.1 & 0.5 & 0.2 \\
{SE} & 0.1 & 0.1 & 0.1 & 0.0 & 0.1 & 0.0 & 0.3 & 0.1 \\
{UC} & 0.4 & 0.0 & 0.1 & 0.1 & 0.1 & 0.0 & 0.4 & 0.1 \\
{UT} & 0.4 & 0.1 & 0.1 & 0.2 & 0.1 & 0.0 & 0.1 & 0.0 \\
{UV} & 0.1 & 0.1 & 0.2 & 0.1 & 0.1 & 0.0 & 0.0 & 0.2 \\
{US} & 0.2 & 0.0 & 0.1 & 0.1 & 0.1 & 0.3 & 0.4 & 0.0 \\
{WK} & 0.1 & 0.1 & 0.1 & 0.1 & 0.1 & 0.1 & 0.5 & 0.1 \\
\bottomrule
\end{tabular}
\end{table*}

\begin{table*}[h!]
\caption{The sample size of temporal neighbors, the number of causal anonymous walks, and the length of input sequences \& the patch size of different methods. The number in parentheses indicates the number of historical interactions used to compute each temporal node embedding during the test stage. For example, on the CP dataset, setting the input sequence length to 2048 denotes that, on average, 99.98\% of historical interactions are used to compute a single temporal node embedding.}
\label{tab:hyper_config_temporal_neighbor_sampling_size}

\centering
\scriptsize
\begin{tabular}{c|ccccccc}
\toprule
\textbf{Datasets} & \textbf{JODIE}  & \textbf{TGAT} & \textbf{TGN} & \textbf{CAWN} & \textbf{TCL} & \textbf{GraphMixer} & \textbf{DyGFormer} \\
\midrule
CP & 10 & 20 & 10 & 128 & 20 & 20 & 2048 (99.98\%) \& 64 \\
CO & 10 & 20 & 10 & 64 & 20 & 20 & 32 (0.49\%) \& 1 \\
EN & 10 & 20 & 10 & 32 & 20 & 20 & 256 (31.69\%) \& 8 \\
FL & 10 & 20 & 10 & 64 & 20 & 20 & 256 (34.93\%) \& 8 \\
{LA} & 10 & 20 & 10 & 128 & 20 & 10 & 512 (41.23\%) \& 16 \\
{MO} & 10 & 20 & 10 & 64 & 20 & 20 & 256 (57.59\%) \& 8 \\
{RE} & 10 & 20 & 10 & 32 & 20 & 10 & 64 (40.48\%)\& 2 \\
SE & 10 & 20 & 10 & 64 & 20 & 20 & 32 (0.05\%) \& 1 \\
UC & 10 & 20 & 10 & 64 & 20 & 20 & 32 (34.13\%) \& 1 \\
UT & 10 & 20 & 10 & 64 & 20 & 20 & 256 (7.05\%) \& 8 \\
UV & 10 & 20 & 10 & 64 & 20 & 20 & 128 (1.66\%) \& 4 \\
US & 10 & 20 & 10 & 32 & 20 & 20 & 256 (63.03\%) \& 8 \\
{WK} & 10 & 20 & 10 & 32 & 20 & 30 & 32 (39.37\%) \& 1 \\
\bottomrule
\end{tabular}
\end{table*}

\begin{table*}[h!]
\caption{Strategies for sampling temporal neighbors during testing and the best-performing variants of EdgeBank.}
\label{tab:hyper_config_smaple_strategy}

\centering
\scriptsize
\begin{tabular}{c|cccccc}
\toprule
\textbf{Datasets} & \textbf{DyRep}  & \textbf{TGAT} & \textbf{TGN} & \textbf{TCL} & \textbf{GraphMixer}  & \textbf{EdgeBank Variant}\\
\midrule
CP          & uniform           & uniform           & uniform           & uniform           & uniform   & EdgeBank$_{tw-ts}$         \\ 
CO             & recent            & recent            & recent            & recent            & recent    & EdgeBank$_{tw-re}$         \\ 
EN               & recent            & recent            & recent            & recent            & recent    & EdgeBank$_{tw-ts}$        \\ 
FL             & recent            & recent            & recent            & recent            & recent    & EdgeBank$_{\infty}$         \\ 
LA              & recent            & recent            & recent            & recent            & recent   & EdgeBank$_{tw-ts}$        \\ 
MO                & recent            & recent            & recent            & recent            & recent   & EdgeBank$_{tw-ts}$        \\ 
RE              & recent            & uniform           & recent            & uniform           & recent   & EdgeBank$_{\infty}$      \\ 
SE         & recent            & recent            & recent            & recent            & recent    & EdgeBank$_{th}$        \\ 
UC                 & recent            & recent            & recent            & recent            & recent    & EdgeBank$_{\infty}$        \\ 
UT            & recent            & uniform           & recent            & uniform           & uniform   & EdgeBank$_{tw-re}$         \\ 
UV             & recent            & recent            & uniform           & uniform           & uniform   & EdgeBank$_{tw-re}$        \\ 
US           & recent            & recent            & recent            & uniform           & recent    & EdgeBank$_{tw-ts}$          \\ 
WK           & recent            & recent            & recent            & recent            & recent   & EdgeBank$_{\infty}$         \\ 
\bottomrule
\end{tabular}

\end{table*}

\subsection{Skewness of Temporal Difference}
\label{appendix:dataset_skewness}
Table~\ref{tab:skewness_of_all_datasets_appendix} presents the skewness of temporal difference for both the baseline (Original TE) and the proposed (\lte{}) time encoding functions. The skewness score is computed using Fisher’s moment coefficient of skewness~\cite{skewness1,skewness2}.  A positive skewness score indicates that the temporal differences are right-skewed, a negative score indicates they are left-skewed, and a score of 0 suggests a balanced distribution. Compared to the baseline encoding function, the proposed \lte{} method effectively reduces the skewness of temporal difference.

\begin{table*}[h!]
\caption{Skewness of temporal difference in different time encoding functions on 13 datasets. {The proposed \lte{} time encoding function effectively reduces the skewness of temporal difference}.}
\label{tab:skewness_of_all_datasets_appendix}
\centering
\scriptsize

\begin{tabular}{c|ccccccccccccc}
\toprule
\textbf{Datasets} & \textbf{CP} & \textbf{CO} & \textbf{EN} & \textbf{FL} & \textbf{LA} & \textbf{MO} & \textbf{RE} & \textbf{SE} & \textbf{UC} & \textbf{UT} & \textbf{UV} & \textbf{US} & \textbf{WK} \\
\midrule
\textbf{Original TE} & 1.73 & 28.81 & 6.35 & 7.45 & 32.76 & 4.14 & 3.47 & 140.42 & 2.38 & 0.34 & 0.62 & 9.22 & 2.64 \\
\midrule
\textbf{\lte{}} & 0.69 & 1.96 & -0.60 & 2.37 & -0.61 & 0.30 & -0.31 & 0.07 & -1.14 & -0.96 & -0.95 & 9.22 & -0.8865 \\
\bottomrule
\end{tabular}
\vskip -0.1in
\end{table*}

\setlength{\tabcolsep}{2.75pt}
\begin{table*}[t]
\centering
\scriptsize
\setlength\cmidrulewidth{\heavyrulewidth}
\caption{AP for transductive link prediction. Negative edges are generated using the random (rnd), historical (hist), and inductive (ind) negative sampling strategies proposed in~\cite{edgebank}. NSS stands for negative sampling strategies. Standard deviations are summarized in Table~\ref{tab:std_main_table}.}

\label{tab:appendix_full_main_results}
\begin{tabular}{c|c|ccccccccccccc}

\toprule
\textbf{NSS} & \textbf{Methods} & \textbf{CP} & \textbf{CO} & \textbf{EN} & \textbf{FL} & \textbf{LA} & \textbf{MO} & \textbf{RE} & \textbf{SE} & \textbf{UC} & \textbf{UT} & \textbf{UV} & \textbf{US} & \textbf{WK} \\
\midrule

\multirow{14}{*}{rnd} & JODIE      & 69.26 & 95.31 & 84.77 & 95.60 & 70.85 & 80.23 & 98.31 & 89.89 & 89.43 & 64.94 & \underline{63.91} & 75.05 & 96.50 \\

& DyRep      & 66.54 & 95.98 & 82.38 & 95.29 & 71.92 & 81.97 & 98.22 & 88.87 & 65.14 & 63.21 & 62.81 & \underline{75.34} & 94.86 \\

& TGAT       & 70.73 & 96.28 & 71.12 & 94.03 & 73.42 & 85.84 & 98.52 & 93.16 & 79.63 & 61.47 & 52.21 & 68.52 & 96.94 \\

& TGN        & 70.88 & 96.89 & 86.53 & 97.95 & 77.07 & \textbf{89.15} & 98.63 & 93.57 & 92.34 & 65.03 & \textbf{65.72} & \textbf{75.99} & 98.45 \\

& CAWN       & 69.82 & 90.26 & 89.56 & 98.51 & 86.99 & 80.15 & 99.11 & 84.96 & 95.18 & \underline{65.39} & 52.84 & 70.58 & 98.76 \\

& EdgeBank   & 64.55 & 92.58 & 83.53 & 89.35 & 79.29 & 57.97 & 94.86 & 74.95 & 76.20 & 60.41 & 58.49 & 58.39 & 90.37 \\

& TCL        & 68.67 & 92.44 & 79.70 & 91.23 & 67.27 & 82.38 & 97.53 & 93.13 & 89.57 & 62.21 & 51.90 & 69.59 & 96.47 \\

& GraphMixer & 75.90 & 91.94 & 82.26 & 90.98 & 75.56 & 82.83 & 97.33 & 93.34 & 93.38 & 62.61 & 52.20 & 71.55 & 97.23 \\

& DyGFormer & \underline{97.91} & \underline{98.31} & \underline{92.46} & \underline{98.92} & \underline{93.01} & {87.66} & \underline{99.22} & \underline{94.66} & 95.66 & 65.07 & 55.88 & 70.44 & \underline{99.02} \\

\cmidrule{2-15}
& \multicolumn{13}{l}{\textbf{with \n{}}} \\
\cmidrule{2-15}

& {GraphMixer} & 78.38 & 95.26 & 90.97 & 96.75 & 88.13 & 83.53 & 98.84 & 93.41 & \underline{96.20} & 62.98 & 57.74 & 71.57 & 98.89 \\

& Imp. (\%) & {\color{darkgreen}3.27\%} & {\color{darkgreen}3.61\%} & {\color{darkgreen}10.59\%} & {\color{darkgreen}6.34\%} & {\color{darkgreen}16.64\%} & {\color{darkgreen}0.85\%} & {\color{darkgreen}1.56\%} & {\color{darkgreen}0.07\%} & {\color{darkgreen}3.02\%} & {\color{darkgreen}0.59\%} & {\color{darkgreen}10.61\%} & {\color{darkgreen}0.03\%} & {\color{darkgreen}1.71\%} \\

& {DyGFormer} & \textbf{98.67} & \textbf{98.70} & \textbf{92.66} & \textbf{98.94} & \textbf{94.03} & \underline{88.49} & \textbf{99.29} & \textbf{94.74} & \textbf{96.72} & \textbf{66.39} & 56.02 & {71.40} & \textbf{99.25} \\ 

& Imp. (\%) & {\color{darkgreen}0.78\%} & {\color{darkgreen}0.40\%} & {\color{darkgreen}0.22\%} & {\color{darkgreen}0.02\%} & {\color{darkgreen}1.10\%} & {\color{darkgreen}0.95\%} & {\color{darkgreen}0.07\%} & {\color{darkgreen}0.08\%} & {\color{darkgreen}1.11\%} & {\color{darkgreen}2.03\%} & {\color{darkgreen}0.25\%} & {\color{darkgreen}1.36\%} & {\color{darkgreen}0.23\%} \\

\midrule

\multirow{14}{*}{hist} & JODIE     & 51.79 & 95.31 & 69.85 & 66.48 & 74.35 & 78.94 & 80.03 & 87.44 & 75.24 & 61.39 & 70.02 & 51.71 & 83.01 \\
& DyRep     & 63.31 & 96.39 & 71.19 & 67.61 & 74.92 & 75.60 & 79.83 & 93.29 & 55.10 & 59.19 & 69.30 & \textbf{86.88} & 79.93 \\
& TGAT      & 67.13 & 96.05 & 64.07 & \underline{72.38} & 71.59 & 82.19 & 79.55 & 95.01 & 68.27 & 55.74 & 52.96 & 62.14 & 87.38 \\
& TGN       & 68.42 & 93.05 & 73.91 & 66.70 & 76.87 & \textbf{87.06} & 81.22 & 94.45 & 80.43 & 58.44 & 69.37 & 74.00 & 86.86 \\
& CAWN      & 66.53 & 84.16 & 64.73 & 64.72 & 69.86 & 74.05 & 80.82 & 85.53 & 65.30 & 55.71 & 51.26 & 68.82 & 71.21 \\
& EdgeBank  & 63.84 & 88.81 & 76.53 & 70.53 & 73.03 & 60.71 & 73.59 & 80.57 & 65.50 & \textbf{81.32} & \textbf{84.89} & 63.20 & 73.35 \\
& TCL       & 65.93 & 93.86 & 70.66 & 70.68 & 59.30 & 77.06 & 77.14 & 94.74 & 80.25 & 55.90 & 52.30 & 80.53 & 89.05 \\
& GraphMixer& 74.34 & 93.29 & 77.98 & 71.47 & 72.47 & 77.77 & 78.44 & 94.93 & 84.11 & 57.05 & 51.20 & 81.65 & \underline{90.90} \\
& DyGFormer & \underline{97.00} & \underline{97.57} & 75.63 & 66.59 & \underline{81.57} & 85.85 & 81.57 & \underline{97.38} & 82.17 & 64.41 & 60.84 & 85.30 & 82.23 \\

\cmidrule{2-15}
& \multicolumn{13}{l}{\textbf{with \n{}}} \\
\cmidrule{2-15}

& GraphMixer & 78.81 & 93.30 & {81.68} & {73.01} & 80.23 & 83.61 & {82.56} & 96.80 & {87.69} & {69.74} & 70.90 & 84.56 & {90.97} \\
& Imp. (\%)       & {\color{darkgreen}6.01\%} & {\color{darkgreen}0.01\%} & {\color{darkgreen}4.74\%} & {\color{darkgreen}2.15\%} & {\color{darkgreen}10.71\%} & {\color{darkgreen}7.51\%} & {\color{darkgreen}5.25\%} & {\color{darkgreen}1.97\%} & {\color{darkgreen}4.26\%} & {\color{darkgreen}22.24\%} & {\color{darkgreen}38.48\%} & {\color{darkgreen}3.56\%} & {\color{darkgreen}0.08\%} \\

& DyGFormer  & {98.96} & {97.72} & {81.02} & 67.77 & {83.40} & {86.26} & {85.18} & {97.56} & {85.89} & 65.16 & {81.72} & {86.10} & 82.38 \\
& Imp. (\%)       & {\color{darkgreen}2.02\%} & {\color{darkgreen}0.15\%} & {\color{darkgreen}7.13\%} & {\color{darkgreen}1.77\%} & {\color{darkgreen}2.24\%} & {\color{darkgreen}0.48\%} & {\color{darkgreen}4.43\%} & {\color{darkgreen}0.18\%} & {\color{darkgreen}4.53\%} & {\color{darkgreen}1.16\%} & {\color{darkgreen}34.32\%} & {\color{darkgreen}0.94\%} & {\color{darkgreen}0.18\%} \\

\midrule

\multirow{14}{*}{ind} & JODIE & 48.42 & 93.43 & 68.96 & 69.07 & 62.67 & 65.23 & 86.96 & 89.82 & 65.99 & 60.42 & 67.79 & 50.27 & 75.65 \\
& DyRep & 58.61 & 94.18 & 67.79 & 70.57 & 64.41 & 61.66 & 86.30 & 93.28 & 54.79 & 60.19 & 67.53 & \textbf{83.44} & 70.21 \\

& TGAT & 68.82 & 94.35 & 63.94 & 75.48 & 71.13 & 75.95 & 89.59 & 94.84 & 68.67 & 60.61 & 52.89 & 61.91 & 87.00 \\

& TGN & 65.34 & 90.18 & 70.89 & 71.09 & 65.95 & 77.50 & 88.10 & 95.13 & 70.94 & 61.04 & 67.63 & 67.57 & 85.62 \\

& CAWN & 67.75 & 89.31 & 75.15 & 69.18 & 67.48 & 73.51 & 91.67 & 88.32 & 64.61 & 62.54 & 52.19 & 65.81 & 74.06 \\

& EdgeBank & 62.16 & 85.20 & 73.89 & \underline{81.08} & \underline{75.49} & 49.43 & 85.48 & 83.69 & 57.43 & \underline{72.97} & 66.30 & 64.74 & 80.63 \\

& TCL & 65.85 & 91.35 & 71.29 & 74.62 & 58.21 & 74.65 & 87.45 & 94.90 & 76.01 & 61.06 & 50.62 & 78.15 & 86.76 \\
& GraphMixer & 69.48 & 90.87 & 75.01 & 74.87 & 68.12 & 74.26 & 85.26 & 94.72 & 80.10 & 60.15 & 51.60 & 79.63 & \underline{88.59} \\

& DyGFormer & \underline{95.44} & 94.75 & 77.41 & 70.92 & 73.97 & \underline{81.24} & 91.11 & \underline{97.68} & 72.25 & 55.79 & 51.91 & 81.25 & 78.29 \\

\cmidrule{2-15}
& \multicolumn{13}{l}{\textbf{with \n{}}} \\
\cmidrule{2-15}

& GraphMixer & 70.94 & {96.12} & {88.95} & {93.64} & {91.06} & 79.82 & {96.19} & 96.09 & {84.12} & {87.73} & {79.53} & {83.31} & {93.89} \\

& Imp. (\%) & {\color{darkgreen}2.10\%} & {\color{darkgreen}5.78\%} & {\color{darkgreen}18.58\%} & {\color{darkgreen}25.07\%} & {\color{darkgreen}33.68\%} & {\color{darkgreen}7.49\%} & {\color{darkgreen}12.82\%} & {\color{darkgreen}1.45\%} & {\color{darkgreen}5.02\%} & {\color{darkgreen}45.85\%} & {\color{darkgreen}54.13\%} & {\color{darkgreen}4.62\%} & {\color{darkgreen}5.98\%} \\

& DyGFormer & {97.25} & {98.47} & {86.23} & 75.55 & 74.03 & {92.39} & {94.37} & {97.76} & {80.13} & 68.01 & {78.19} & {81.31} & 78.96 \\

& Imp. (\%) & {\color{darkgreen}1.90\%} & {\color{darkgreen}3.93\%} & {\color{darkgreen}11.39\%} & {\color{darkgreen}6.53\%} & {\color{darkgreen}0.08\%} & {\color{darkgreen}13.72\%} & {\color{darkgreen}3.58\%} & {\color{darkgreen}0.08\%} & {\color{darkgreen}10.91\%} & {\color{darkgreen}21.90\%} & {\color{darkgreen}50.63\%} & {\color{darkgreen}0.07\%} & {\color{darkgreen}0.86\%} \\

\bottomrule
\end{tabular}
\end{table*}

\section{Comprehensive Results for Transductive Temporal Link Prediction}
\label{sec:appendix_transductive_results}
This section presents additional results for transductive link prediction. Section~\ref{sec:appendix_comp_table_1_results} presents the comprehensive results of Table~\ref{tab:main_result}. Section~\ref{sec:appendix_transductive_port} shows that applying \lte{} and \trc{} individually also improves the link prediction accuracy of the integrated model. Section~\ref{sec:when_log_te_work} discusses how the effectiveness of \lte{} is influenced by the skewness of temporal differences in a dataset and the configurations of underlying TGNNs. Section~\ref{sec:exp_hyper_param_gamma} explores the impact of the hyperparameter $\gamma$ in our \trc{} module. Section~\ref{sec:exp_lha_k_and_aggregation_function} studies the robustness of \n{} under different aggregation strategies and values of $k$ in our \trc{} module. Section~\ref{sec:appendix_trc_JODIE_forget} highlights that TGNNs suffer from the loss of interaction histories, which reduces their link prediction performance. Our \trc{} module helps mitigate this loss and improves TGNNs' performance. Section~\ref{sec:appendix_trc_code_start} evaluates the effectiveness of \trc{} under conditions of limited memory.

\subsection{Comprehensive Results of Table~\ref{tab:main_result}}
\label{sec:appendix_comp_table_1_results}
Table~\ref{tab:appendix_full_main_results} presents the comprehensive results of Table~\ref{tab:main_result}. Results show that \n{} consistently improves the link prediction accuracy of underlying TGNNs and remains effective under different negative sampling strategies.

\subsection{Adaptivity of \lte{} and \trc{} to Various Types of TGNNs}
\label{sec:appendix_transductive_port}
Table~\ref{tab:portability_appendix} demonstrates the comprehensive results of Table~\ref{tab:portability_main_text} in Section~\ref{sec:portability_main_text}. It also presents the link prediction performance of TGNNs integrated with the proposed \lte{} and \trc{}, alongside their vanilla counterparts. The results suggest that \n{} is adaptable to different types of TGNNs, consistently improving the performance of underlying models. Moreover, applying \lte{} and \trc{} individually also improves the link prediction accuracy of the integrated model.


\begin{table*}[t]
\caption{AP for transductive setting. The proposed \lte{} and \trc{} can be integrated into various types of TGNNs. When applied individually, both \lte{} and \trc{} improve the link prediction accuracy of the integrated models. No \lte{} and \n{} results are reported for JODIE because it does not utilize the time encoding function.}
\label{tab:portability_appendix}

\centering
\scriptsize

\begin{tabular}{c|cccc}
\toprule
\textbf{Method} & \textbf{EN} & \textbf{LA} & \textbf{UC} & \textbf{UV} \\
\midrule
    TGN  & 87.09 $\pm$ 1.04 & 75.73 $\pm$ 1.53 & 91.87 $\pm$ 1.42 & 65.68 $\pm$ 1.19 \\
    TGAT & 72.89 $\pm$ 1.12 & 73.37 $\pm$ 0.07 & 79.29 $\pm$ 0.06 & 53.14 $\pm$ 0.56 \\
    CAWN  & 88.39 $\pm$ 0.07 & 86.99 $\pm$ 0.01 & 94.97 $\pm$ 0.08 & 52.88 $\pm$ 0.08 \\
    JODIE & 84.79 $\pm$ 4.80 & 70.17 $\pm$ 3.82 & 89.15 $\pm$ 0.98 & 63.52 $\pm$ 0.10 \\

\midrule
\multicolumn{5}{l}{\textbf{w/ \lte{}}} \\
\midrule

    TGN &  90.23 $\pm$ 0.47 & 85.03 $\pm$ 1.45 & 94.17 $\pm$ 1.12 & 64.90 $\pm$ 0.32 \\
    \multicolumn{1}{r|}{Imp. (\%)}  & {\color{darkgreen}3.61\%} & {\color{darkgreen}12.27\%} & {\color{darkgreen}2.51\%} & {\color{RawSienna}-1.20\%} \\
    \hline
    
    TGAT & 84.75 $\pm$ 0.04 & 83.49 $\pm$ 0.01 & 94.85 $\pm$ 0.50 & 54.97 $\pm$ 3.96 \\
    \multicolumn{1}{r|}{Imp. (\%)} & {\color{darkgreen}16.27\%} & {\color{darkgreen}13.79\%} & {\color{darkgreen}19.62\%} & {\color{darkgreen}3.45\%} \\
    \hline
    
    CAWN  & 90.70 $\pm$ 0.39 & 88.76 $\pm$ 0.03 & 96.63 $\pm$ 0.08 & 52.92 $\pm$ 0.77 \\
    \multicolumn{1}{r|}{Imp. (\%)}  & {\color{darkgreen}2.61\%} & {\color{darkgreen}2.03\%} & {\color{darkgreen}1.75\%} & {\color{darkgreen}0.08\%} \\
        
\midrule
\multicolumn{5}{l}{\textbf{w/ \trc{}}} \\ 
\midrule

TGN  & 90.20 $\pm$ 0.60 & 90.12 $\pm$ 0.30 & 93.34 $\pm$ 0.55 & 67.95 $\pm$ 1.22 \\
\multicolumn{1}{r|}{Imp. (\%)}  & {\color{darkgreen}3.57\%} & {\color{darkgreen}19.00\%} & {\color{darkgreen}1.61\%} & {\color{darkgreen}3.46\%} \\
\hline

TGAT  & 88.06 $\pm$ 0.12 & 87.97 $\pm$ 0.23 & 89.74 $\pm$ 0.11 & 57.58 $\pm$ 0.27 \\
\multicolumn{1}{r|}{Imp. (\%)}  & {\color{darkgreen}20.81\%} & {\color{darkgreen}19.89\%} & {\color{darkgreen}13.17\%} & {\color{darkgreen}8.37\%} \\
\hline

CAWN  & 90.13 $\pm$ 0.01 & 89.29 $\pm$ 0.04 & 95.16 $\pm$ 0.03 & 56.50 $\pm$ 0.54 \\
\multicolumn{1}{r|}{Imp. (\%)}  & {\color{darkgreen}1.96\%} & {\color{darkgreen}2.64\%} & {\color{darkgreen}0.21\%} & {\color{darkgreen}6.85\%} \\
\hline

JODIE  & 90.62 $\pm$ 0.01 & 87.95 $\pm$ 0.01 & 92.44 $\pm$ 0.02 & 65.57 $\pm$ 0.34 \\
\multicolumn{1}{r|}{Imp. (\%)}  & {\color{darkgreen}6.88\%} & {\color{darkgreen}25.33\%} & {\color{darkgreen}3.68\%} & {\color{darkgreen}3.22\%} \\

\midrule
\multicolumn{5}{l}{\textbf{w/ \n{}}} \\ 
\midrule

{TGN}  & 92.34 $\pm$ 0.04 & 92.83 $\pm$ 0.01 & 95.36 $\pm$ 0.11 & 67.80 $\pm$ 0.59 \\
\multicolumn{1}{r|}{Imp. (\%)}  & {\color{darkgreen}6.03\%} & {\color{darkgreen}22.58\%} & {\color{darkgreen}3.80\%} & {\color{darkgreen}3.22\%} \\
\hline

{TGAT}  & 91.37 $\pm$ 0.04 & 91.60  $\pm$ 0.13 & 96.36 $\pm$ 0.18 & 60.03  $\pm$ 0.34\\
\multicolumn{1}{r|}{Imp. (\%)}  & {\color{darkgreen}25.35\%} & {\color{darkgreen}24.85\%} & {\color{darkgreen}21.53\%} & {\color{darkgreen}12.98\%} \\
\hline

{CAWN}  & 91.23 $\pm$ 0.06 & 91.02 $\pm$ 0.00 & 96.69 $\pm$ 0.13 & 57.49 $\pm$ 0.18\\
\multicolumn{1}{r|}{Imp. (\%)} & {\color{darkgreen}3.21\%} & {\color{darkgreen}4.64\%} & {\color{darkgreen}1.81\%} & {\color{darkgreen}8.72\%}  \\

\bottomrule

\end{tabular}
\end{table*}


\begin{table*}[b]
\caption{AP of GraphMixer with varying values of the hyperparameter $\gamma$ when integrated with our \trc{} module. The first and the second best performances are marked in \textbf{bold} and \underline{underlined} respectively.}
\label{tab:hyper_gamma}

\centering
\scriptsize

\begin{tabular}{ccccc}
\toprule
 & \textbf{EN} & \textbf{LA} & \textbf{UC} & \textbf{UV} \\

\midrule
GraphMixer & 82.26 & 75.56 & 93.38 & 52.20 \\ 

\midrule
\multicolumn{5}{l}{\textbf{with \n{}} } \\
\midrule
$\gamma=0.1$               & 90.41 & 85.97 & 95.80 & \textbf{58.02} \\
$\gamma=0.3$               & 90.87 & 87.42 & 96.11 & \underline{57.99} \\
$\gamma=0.5$               & \textbf{90.97} & 87.94 & \underline{96.22} & 57.95 \\
$\gamma=0.7$               & \underline{90.96} & \underline{88.18} & \textbf{96.24} & 57.87 \\
$\gamma=0.9$               & 90.77 & \textbf{88.19} & 96.20 & 57.74 \\
$\gamma=1$                 & 90.60 & 88.09 & 96.16 & 57.55 \\
\bottomrule
\end{tabular}
\end{table*}

\subsection{Application Scenarios of \lte{}}
\label{sec:when_log_te_work}
The effectiveness of the proposed \lte{} is influenced by the distribution of temporal distances within the dataset. Based on the skewness of the temporal differences in the original TE, we classify the datasets into two types: (1) skewed, where small temporal differences dominate, while a considerable number of large temporal differences exist; and (2) balanced, where the temporal differences roughly follow a normal distribution. As shown in Table~\ref{tab:ablation} and Table~\ref{tab:skewness_of_all_datasets_appendix}, for balanced datasets (e.g., UV and UT), the improvements observed in GraphMixer are marginal. This is because \lte{} is specifically designed to address the skewness of temporal differences, which is rarely presented in balanced datasets. For datasets with skewed temporal differences (e.g., CP, EN, and UC), \lte{} significantly improves GraphMixer’s performance compared to the vanilla version with the original TE.

\subsection{Analysis of the Hyperparameter $\gamma$ in \trc{}}
\label{sec:exp_hyper_param_gamma}
In this experiment, we study the impact of the hyperparameter $\gamma$ in our \trc{} module on link prediction performance. We train, validate, and test models with $\gamma = (0.1, 0.3, 0.5, 0.7, 0.9, 1)$ and report the test AP in Table~\ref{tab:hyper_gamma}. As shown, integrating GraphMixer into \n{} consistently improves its performance over its vanilla counterpart across all values of $\gamma$. In addition, $\gamma$ affects link prediction accuracy differently across different datasets. For the UV dataset, the performance of GraphMixer with \trc{} remains similar across different values of $\gamma$. When it comes to the EN, LA, and UC datasets, higher $\gamma$ values generally yield better performance. This is because larger $\gamma$ values place more weight on recent historical interactions, while reducing the influence of older ones. These results suggest that prioritizing recent interactions is more advantageous for these datasets.

\subsection{Analysis of the Choice of Aggregation Strategies and the Value of $k$ in \trc{}}
\label{sec:exp_lha_k_and_aggregation_function}
To predict the future link between two nodes, \trc{} retrieves all stored $k$ historical edge embeddings for the target node pair and summarizes them into a single vector using an aggregation function. In this experiment, we study the effectiveness of \n{} under different values of $k$ and aggregation strategies. By default, we set $k=1$ and use the \textbf{most-recent} aggregator. Note that our \trc{} module uses an exponentially weighted moving average to compute the dedicated historical edge embeddings for a given target pair of nodes, as shown in Equation~\ref{eq:trc_context_vector}. Thus, even if $k=1$, \trc{} considers all the interaction history between the pair of nodes.

In Table~\ref{tab:aggregation_function_and_k}, we show the performance of \n{} with different $k$ values and aggregation functions. The \textbf{mean} aggregator computes the average of the $k$ historical edge embeddings, whereas the \textbf{max} aggregator follows the implementation in GraphSAGE~\cite{graphsage}, where each historical edge embedding is passed through a fully connected neural network, followed by an element-wise max-pooling operation across all transformed vectors. Overall, the performance of different variants of \n{} is better than the vanilla version (without \n{}), indicating the effectiveness of our design. It is also important to note that larger $k$ value does not necessarily lead to better performance. This is expected as different datasets may have different levels of dependency on the interaction history for temporal link prediction. In addition, \n{} is not sensitive to the choice of aggregation functions.

\begin{table}[h]
    \centering
    \caption{Test AP of DyGFormer on the EN and CP datasets with different $k$ values and aggregation strategies.}
    \begin{tabular}{c|cc}
        \toprule
        \textbf{Method} &  \textbf{EN} & \textbf{CP} \\
        \midrule
        DyGFormer (Vanilla) &  92.46  & 97.91 \\
        \midrule
        with \n{} \\
        most-recent &  92.66  & 98.67 \\
        \midrule
        mean ($k=1$) & 92.66  & 98.67\\
        mean ($k=2$) &  92.60  & 98.76 \\
        mean ($k=3$) & 92.56  & \textbf{98.77}\\
        \midrule
        max ($k=1$) & 92.66  & 98.67\\
        max ($k=2$) & \textbf{92.84} & 98.74  \\
        max ($k=3$) & 92.60 & 98.74\\
        \bottomrule
    \end{tabular}
    \label{tab:aggregation_function_and_k}
\end{table}

\subsection{Preventing the Forgetting of Interaction Histories by \trc{}}
\label{sec:appendix_trc_JODIE_forget}
\begin{figure}[t!]

\begin{center}
\centerline{\includegraphics[width=\textwidth]{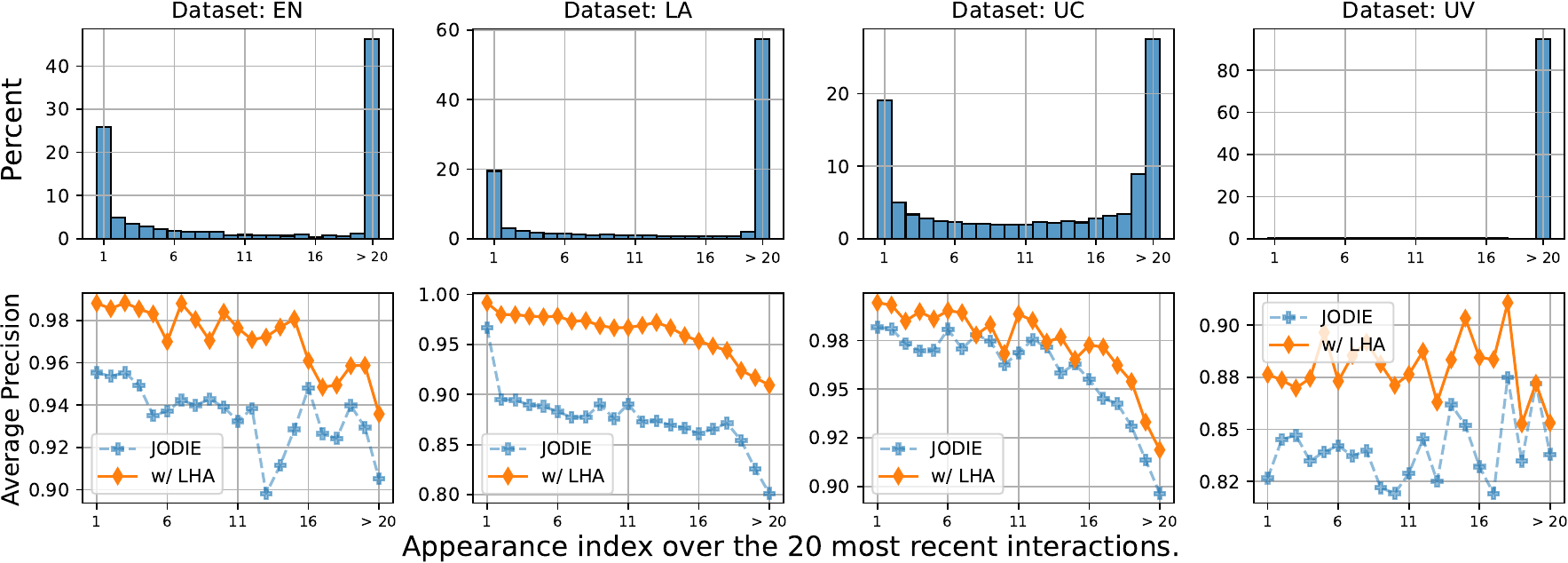}}
\caption{AP of JODIE to the number of interaction histories retained. An increase in the appearance index indicates that fewer interaction histories are retained for predicting future links.}
\label{fig:trc_most_recent_matching_index}
\end{center}
\vskip -0.3in
\end{figure}

In Figure~\ref{fig:trc_direct_match}, we demonstrate that the loss of interaction histories occurs in GraphMixer~\cite{graphmixer} and degrades its link prediction performance. Here we extend our analysis to JODIE~\cite{JODIE}. JODIE adopts two RNNs to maintain an evolving temporal memory for each node and uses stored historical node states to compute temporal node embeddings. When predicting the link between two nodes,  if neither node appears in the other node’s 20 most recent interactions, we consider their interaction histories as no longer retained in their temporal node embeddings. Figure~\ref{fig:trc_most_recent_matching_index} presents the distribution of the appearance index along with the link prediction accuracy. The appearance index represents the earliest position at which at least one node appears in the interaction sequence of the other. For instance, an index value of 1 means that at least one node is the most recent encounter of the other, while a larger index indicates that fewer historical interactions between the nodes are retained. The term ``$> 20$'' denotes the case where neither node appears in the most recent 20 interactions of the other and their interaction histories are forgotten.

As shown in Figure~\ref{fig:trc_most_recent_matching_index}, the ``$> 20$'' term has the highest percentage across all examined datasets. This suggests that the loss of historical interactions also occurs in memory-based TGNNs. On the EN, LA, and UC datasets, the AP of the vanilla JODIE model decreases as the appearance index increases, indicating that retaining fewer historical interactions leads to worse link prediction performance. In comparison, when the proposed \trc{} is incorporated, the AP of JODIE consistently improves across all datasets, demonstrating the effectiveness of our \trc{} module in mitigating the loss of historical interactions.

\begin{figure}[b]

\begin{center}
\centerline{\includegraphics[width=\textwidth]{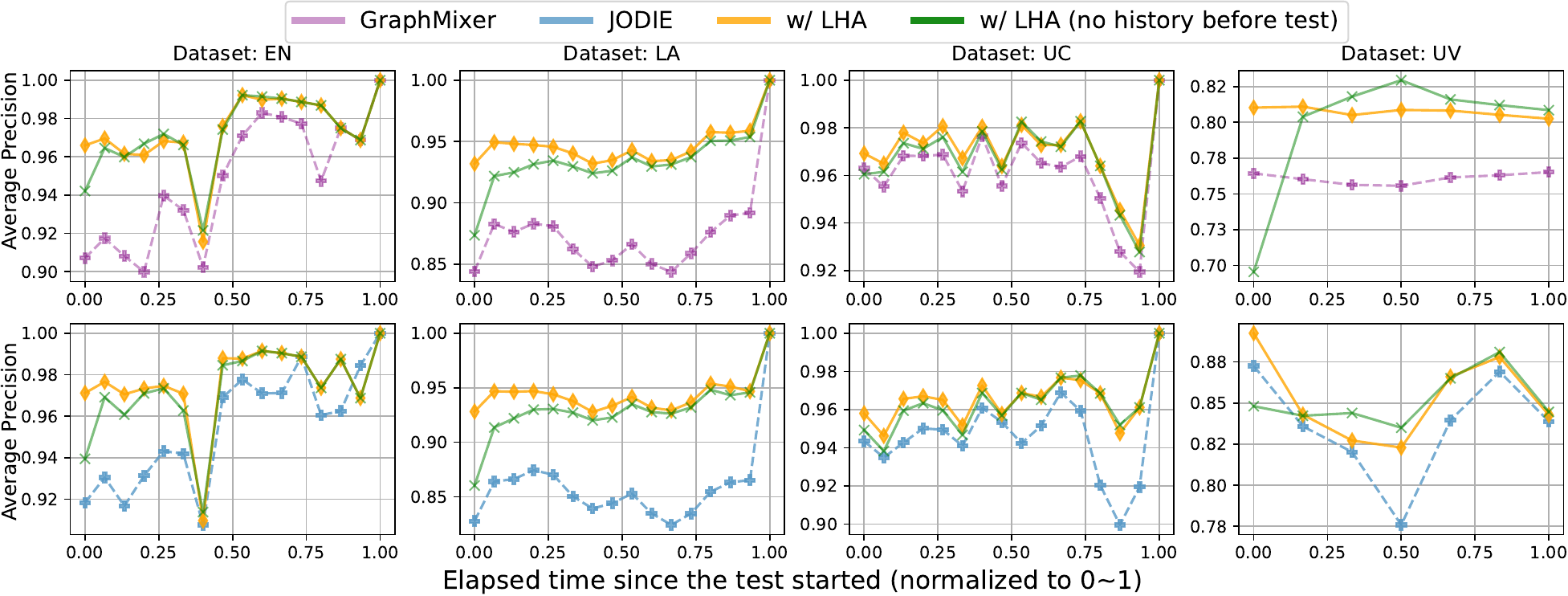}}
\caption{\trc{} consistently improves the performance of TGNNs throughout the entire testing phase. The ``w/ \trc{}'' variant denotes that interaction histories accumulated during the training and validation phases are loaded at the beginning of the test phase. In contrast, the ``w/ \trc{} (no history before test)'' variant starts the test phase without any interaction histories stored in the \trc{} memory module. Test node pairs are grouped into chronological bins, and the average link prediction accuracy for each bin is reported.}
\label{fig:cold_exp_vs_time}
\end{center}
\end{figure}

\subsection{Cold Start of \trc{} in Inference} 
\label{sec:appendix_trc_code_start}
\trc{} maintains the historical connections between nodes. It may happen that in the training set, two nodes never interact. In this experiment, we evaluate how \trc{} performs in that case. Specifically, we introduce a variant called ``w/ \trc{} (no history before test)'', where the interaction histories accumulated during the training and validation phases are not loaded when the test stage begins. In other words, interaction histories must be constructed from scratch, as the test progresses. This variant is compared against the default \trc{} setting, ``w/ \trc{},'' where the history accumulated during training and validation is loaded before testing. Figure~\ref{fig:cold_exp_vs_time} illustrates the test AP of various methods. The x-axis shows the test phase timeline, with timestamps normalized to [0, 1] for better visualization. 

First, integrating \trc{} into GraphMixer and JODIE consistently boosts their performance throughout the testing phase, surpassing their vanilla counterparts. Second, even when \trc{} history is not initialized at the beginning of the test phase (i.e., in the ``w/ \trc{} (no history before test)'' variant), the augmented TGNNs still outperform their vanilla versions on the EN, LA, and UC datasets. This indicates that the \trc{} module provides significant performance improvements, even with a limited history in the early stage of inference. As the test progresses and more interaction histories are stored in the \trc{}, the performance of the ``w/ \trc{} (no history before test)'' variant gradually converges to that of the ``w/ \trc{}'' variant. This is expected, as both variants share increasingly similar histories, leading to similar performance over time.

\section{Comprehensive Results for Inductive Temporal Link Prediction}
\label{appendix:inductive_link_prediction}
This section presents comprehensive results for inductive link prediction. Section~\ref{sec:appendix_inductive_main} demonstrates that \lte{} and \trc{} improve the link prediction performance of integrated models in the inductive setting. Section~\ref{sec:appendix_inductive_ablation} presents the results of an ablation study in the inductive setting, showing that both \lte{} and \trc{} remain effective. In Section~\ref{sec:appendix_inductive_more_negative}, we show that \lte{} and \trc{} are robust to an increasing number of negative links per positive link in the inductive setting. Finally, Section~\ref{sec:appendix_inductive_portability} shows that the proposed \lte{} and \trc{} can be integrated into various types of TGNN while enhancing their performance in the inductive setting.


\begin{table*}[t]
\caption{AP for inductive link prediction. Imp. (\%) denotes the percentage of improvement. Improved results are colored in {\color{darkgreen}blue}. The first and the second best performances are marked in \textbf{bold} and \underline{underlined} respectively. For dataset names, please kindly refer to Section~\ref{appendix:dataset_description}.}
\label{tab:appendix_inductive_main}

\centering
\scriptsize

\begin{tabular}{lccccccc}
\toprule
        \textbf{Methods} & \textbf{CP} & \textbf{CO} & \textbf{EN} & \textbf{FL} & \textbf{LA} & \textbf{MO} & \textbf{RE} \\
\midrule
        JODIE & 53.92 $\pm$ 0.94 & 94.34 $\pm$ 1.45 & 80.67 $\pm$ 2.11 & 94.74 $\pm$ 0.37 & 78.31 $\pm$ 3.82 & 79.63 $\pm$ 1.92 & 96.50 $\pm$ 0.13 \\
        DyRep & 54.02 $\pm$ 0.76 & 92.18 $\pm$ 0.41 & 74.55 $\pm$ 3.95 & 92.88 $\pm$ 0.73 & 83.02 $\pm$ 1.48 & 81.07 $\pm$ 0.44 & 96.09 $\pm$ 0.11 \\
        TGAT & 55.18 $\pm$ 0.79 & 95.87 $\pm$ 0.11 & 67.63 $\pm$ 0.96 & 88.73 $\pm$ 0.33 & 78.48 $\pm$ 0.07 & 85.50 $\pm$ 0.19 & 97.09 $\pm$ 0.04 \\
        TGN & 54.10 $\pm$ 0.93 & 93.82 $\pm$ 0.99 & 78.74 $\pm$ 1.53 & 95.03 $\pm$ 0.60 & 80.05 $\pm$ 4.00 & \textbf{89.04} $\pm$ 1.17 & 97.50 $\pm$ 0.07 \\
        CAWN & 55.80 $\pm$ 0.69 & 89.55 $\pm$ 0.30 & 84.97 $\pm$ 0.21 & 97.60 $\pm$ 0.02 & 89.48 $\pm$ 0.02 & 81.42 $\pm$ 0.24 & 98.62 $\pm$ 0.01 \\
        TCL & 54.30 $\pm$ 0.66 & 91.11 $\pm$ 0.12 & 76.14 $\pm$ 0.79 & 83.41 $\pm$ 0.07 & 73.53 $\pm$ 1.66 & 80.60 $\pm$ 0.22 & 94.09 $\pm$ 0.07 \\
        GraphMixer & 57.46 $\pm$ 0.11 & 90.61 $\pm$ 0.09 & 75.94 $\pm$ 0.21 & 83.00 $\pm$ 0.04 & 82.06 $\pm$ 0.25 & 81.35 $\pm$ 0.01 & 95.23 $\pm$ 0.03 \\
        DyGFormer & \underline{88.14} $\pm$ 0.19 & \underline{98.06} $\pm$ 0.01 & \underline{89.86} $\pm$ 0.28 & \underline{97.79} $\pm$ 0.01 & \underline{94.20} $\pm$ 0.17 & \underline{87.24} $\pm$ 0.35 & \underline{98.83} $\pm$ 0.03 \\
\midrule
        \multicolumn{7}{l}{with \n{}} & \\ 
\midrule
        {GraphMixer} & 58.57 $\pm$ 0.92 & 94.49 $\pm$ 0.21 & 85.54 $\pm$ 0.45 & 92.64 $\pm$ 0.06 & 90.71 $\pm$ 0.64 & 81.55 $\pm$ 0.10 & 97.93 $\pm$ 0.01 \\
        \multicolumn{1}{r}{Imp. (\%)} & {\color{darkgreen}1.93\%} & {\color{darkgreen}4.28\%} & {\color{darkgreen}12.63\%} & {\color{darkgreen}11.62\%} & {\color{darkgreen}10.55\%} & {\color{darkgreen}0.25\%} & {\color{darkgreen}2.84\%} \\
\midrule
        {DyGFormer} & \textbf{88.97} $\pm$ 0.41 & \textbf{98.30} $\pm$ 0.00 & \textbf{89.87} $\pm$ 0.06 & \textbf{97.85} $\pm$ 0.02 & \textbf{95.08} $\pm$ 0.13 & 88.08 $\pm$ 0.23 & \textbf{98.91} $\pm$ 0.00 \\
        \multicolumn{1}{r}{Imp. (\%)} & {\color{darkgreen}0.94\%} & {\color{darkgreen}0.24\%} & {\color{darkgreen}0.01\%} & {\color{darkgreen}0.06\%} & {\color{darkgreen}0.93\%} & {\color{darkgreen}0.96\%} & {\color{darkgreen}0.08\%} \\
\bottomrule
\end{tabular}

\begin{tabular}{cccccccc}
\toprule
        \textbf{Methods} & \textbf{SE} & \textbf{UC} & \textbf{UT} & \textbf{UV} & \textbf{US} & \textbf{WK} \\
\midrule
        JODIE & 91.96 $\pm$ 0.48 & 79.27 $\pm$ 1.97 & 59.65 $\pm$ 0.77 & \textbf{56.86} $\pm$ 0.06 & 54.93 $\pm$ 2.29 & 94.82 $\pm$ 0.20 \\
        DyRep & 90.04 $\pm$ 0.47 & 57.48 $\pm$ 1.87 & 57.02 $\pm$ 0.69 & 53.62 $\pm$ 2.22 & \underline{57.28} $\pm$ 0.71 & 92.43 $\pm$ 0.37 \\
        TGAT & 91.41 $\pm$ 0.16 & 79.11 $\pm$ 0.28 & 61.03 $\pm$ 0.18 & 51.98 $\pm$ 0.14 & 51.00 $\pm$ 3.11 & 96.22 $\pm$ 0.07 \\
        TGN & 90.77 $\pm$ 0.86 & 86.66 $\pm$ 1.83 & 58.31 $\pm$ 3.15 & {54.13} $\pm$ 2.37 & \textbf{58.63} $\pm$ 0.37 & 97.83 $\pm$ 0.04 \\
        CAWN & 79.94 $\pm$ 0.18 & 92.45 $\pm$ 0.06 & \textbf{65.24} $\pm$ 0.21 & 49.29 $\pm$ 0.87 & 53.17 $\pm$ 1.20 & 98.24 $\pm$ 0.03 \\
        TCL & 91.55 $\pm$ 0.09 & 87.36 $\pm$ 2.03 & 62.21 $\pm$ 0.12 & 51.60 $\pm$ 0.97 & 52.59 $\pm$ 0.97 & 96.22 $\pm$ 0.17 \\
        GraphMixer & 91.71 $\pm$ 0.13 & 91.39 $\pm$ 0.05 & 62.26 $\pm$ 0.13 & 50.58 $\pm$ 0.45 & 50.72 $\pm$ 0.64 & 96.47 $\pm$ 0.13 \\
        DyGFormer & \underline{93.16} $\pm$ 0.08 & \underline{94.40} $\pm$ 0.21 & \underline{63.86} $\pm$ 0.35 & 56.16 $\pm$ 0.07 & 54.93 $\pm$ 0.53 & \underline{98.54} $\pm$ 0.07 \\
\midrule
        \multicolumn{6}{l}{with \n{}} & \\
\midrule
        GraphMixer & 93.03 $\pm$ 0.46 & 93.44 $\pm$ 0.29 & 62.40 $\pm$ 1.50 & \underline{56.63} $\pm$ 0.36 & 51.44 $\pm$ 0.75 & 98.17 $\pm$ 0.01 \\
        \multicolumn{1}{r}{Imp. (\%)} & {\color{darkgreen}1.44\%} & {\color{darkgreen}2.24\%} & {\color{darkgreen}0.22\%} & {\color{darkgreen}11.96\%} & {\color{darkgreen}1.42\%} & {\color{darkgreen}1.76\%} \\
\midrule
        DyGFormer  & \textbf{93.17} $\pm$ 0.01 & \textbf{95.67} $\pm$ 0.09 & 64.25 $\pm$ 0.24 & 56.23 $\pm$ 0.17 & {55.08} $\pm$ 0.21 & \textbf{98.88}$\pm$ 0.07 \\
        
        \multicolumn{1}{r}{Imp. (\%)} & {\color{darkgreen}0.01\%} & {\color{darkgreen}1.35\%} & {\color{darkgreen}0.61\%} & {\color{darkgreen}0.12\%} & {\color{darkgreen}0.27\%} & {\color{darkgreen}0.35\%} \\
\bottomrule
\end{tabular}

\end{table*}
\setlength{\tabcolsep}{4pt}

\subsection{Main Results for Inductive Link Prediction}
\label{sec:appendix_inductive_main}
Table~\ref{tab:appendix_inductive_main} summarizes the test performance of methods in the inductive setting. The results demonstrate that \n{} improves GraphMixer and DyGFormer on all the 13 datasets. This highlights the effectiveness and versatility of \n{}, suggesting \n{} can enhance link prediction performance in inductive settings.

\subsection{Ablation Study for Inductive Link Prediction}
\label{sec:appendix_inductive_ablation}
In this experiment, we study the individual contributions of \lte{} and \trc{} to the performance improvements of the integrated TGNNs. Table~\ref{tab:appendix_inductive_ablation} presents the test performance of \n{} and its two variants: \textbf{w/ \lte{}}, where we replace the original TE in TGNNs with the proposed \lte{} and keep the rest unchanged; \textbf{w/ \trc{}}, where we integrate the \trc{} module into TGNNs and keep the rest unchanged.

First, replacing the original time encoding function with the proposed \lte{} improves the performance of GraphMixer on 11 datasets and the performance of DyGFormer on 12 datasets. This suggests that \lte{} can enhance the link prediction performance of integrated models in the inductive setting. Second, incorporating the \trc{} module improves the performance of TGNNs. In the inductive setting, even though no test edges are observed during training (i.e., no historical interactions are stored in the \trc{} memory at the beginning of the test phase), the \trc{} module still improves the performance of GraphMixer and DyGFormer on 12 datasets. This is because \trc{} can update its memory as the test stage progresses and improve the accuracy of subsequent link predictions. These results suggest that the \trc{} unit remains effective in the inductive setting.

\subsection{Robustness to the Increase of Negative Links (Inductive)}
\label{sec:appendix_inductive_more_negative}
In Section~\ref{sec:exp_more_negative}, we demonstrate that \n{} is robust to an increasing ratio of negative links to positive links under the transductive setting. In this experiment, we evaluate the robustness of \n{} in the inductive setting. Table~\ref{tab:appendix_inductive_more_negative} presents the test performance of TGNNs under the inductive setting. The term ``NEG=50'' denotes that the connection probability of each positive link is evaluated against 50 negative links. The results demonstrate that the performance of underlying TGNNs consistently improves across various numbers of negative links per positive link on all examined datasets. These findings suggest that the proposed \n{} framework remains effective against the increasing ratio of negative links in the inductive setting.

\subsection{Adaptivity to Different TGNN Architectures (Inductive)}
\label{sec:appendix_inductive_portability}
In Section~\ref{sec:portability_main_text} and Section~\ref{sec:appendix_transductive_port}, we demonstrate that the proposed \n{}, \lte{}, and \trc{} can enhance the link prediction performance of underlying TGNNs in the transductive setting. In this experiment, we explore whether the proposed methods can also improve the performance of TGNNs in the inductive setting. Table~\ref{tab:appendix_inductive_portability} presents the performance of the vanilla TGNNs along with those enhanced with the proposed modules. No results for w/ \lte{} and w/ \n{} are reported for JODIE, as it does not incorporate the time encoding function. The results demonstrate that \lte{}, \trc{}, and \n{} consistently improve the inductive AP of TGNNs. This suggests \n{} and the two novel modules \lte{} and \trc{} remain effective in the inductive setting and can improve the performance of various TGNNs.

\begin{table*}[t]
\caption{Ablation study. AP for inductive link prediction. For dataset names, please kindly refer to Section~\ref{appendix:dataset_description}.}
\label{tab:appendix_inductive_ablation}
\centering
\scriptsize
\begin{tabular}{c|ccccccc}
\toprule
        \textbf{Methods} & \textbf{CP} & \textbf{CO} & \textbf{EN} & \textbf{FL} & \textbf{LA} & \textbf{MO} & \textbf{RE} \\
\midrule
        GraphMixer & 57.46 $\pm$ 0.11 & 90.61 $\pm$ 0.09 & 75.94 $\pm$ 0.21 & 83.00 $\pm$ 0.04 & 82.06 $\pm$ 0.25 & 81.35 $\pm$ 0.01 & 95.23 $\pm$ 0.03  \\
        w/ \lte{} & 58.25 $\pm$ 1.14 & 90.83 $\pm$ 0.27 & 76.15 $\pm$ 0.26 & 83.05 $\pm$ 0.18 & 81.22 $\pm$ 0.90 & 81.73 $\pm$ 0.18 & 95.31 $\pm$ 0.01 \\
        Imp. (\%)  & {\color{darkgreen}1.37\%} & {\color{darkgreen}0.25\%} & {\color{darkgreen}0.27\%} & {\color{darkgreen}0.07\%} & {\color{RawSienna}-1.02\%} & {\color{darkgreen}0.47\%} & {\color{darkgreen}0.08\%} \\
\midrule
        w/ \trc{}  & 57.30 $\pm$ 0.30 & 95.03 $\pm$ 1.06 & 85.16 $\pm$ 0.15 & 92.60 $\pm$ 0.06 & 90.92 $\pm$ 0.24 & 81.50 $\pm$ 0.06 & 97.94 $\pm$ 0.01  \\
        Imp. (\%)  & {\color{RawSienna}-0.29\%} & {\color{darkgreen}4.88\%} & {\color{darkgreen}12.13\%} & {\color{darkgreen}11.57\%} & {\color{darkgreen}10.80\%} & {\color{darkgreen}0.19\%} & {\color{darkgreen}2.85\%} \\ 
\midrule
        w/ \n{} & 58.57 $\pm$ 0.92 & 94.49 $\pm$ 0.21 & 85.54 $\pm$ 0.45 & 92.64 $\pm$ 0.06 & 90.71 $\pm$ 0.64 & 81.55 $\pm$ 0.10 & 97.93 $\pm$ 0.01  \\
        Imp. (\%)  & {\color{darkgreen}1.93\%} & {\color{darkgreen}4.28\%} & {\color{darkgreen}12.63\%} & {\color{darkgreen}11.62\%} & {\color{darkgreen}10.55\%} & {\color{darkgreen}0.25\%} & {\color{darkgreen}2.84\%} \\ 
\midrule
\noalign{\vskip 0.05in}
\midrule
        DyGFormer & 88.14 $\pm$ 0.19 & 98.06 $\pm$ 0.01 & 89.86 $\pm$ 0.28 & 97.79 $\pm$ 0.01 & 94.20 $\pm$ 0.17 & 87.24 $\pm$ 0.35 & 98.83 $\pm$ 0.03 \\
        w/ \lte{} & 88.72 $\pm$ 0.36 & 98.10 $\pm$ 0.01 & 89.97 $\pm$ 0.51 & 97.82 $\pm$ 0.02 & 94.94 $\pm$ 0.04 & 88.30 $\pm$ 0.28 & 98.93 $\pm$ 0.03 \\
        Imp. (\%) & {\color{darkgreen}0.66\%} & {\color{darkgreen}0.04\%}
        & {\color{darkgreen}0.12\%} & {\color{darkgreen}0.03\%} & {\color{darkgreen}0.79\%} & {\color{darkgreen}1.21\%} & {\color{darkgreen}0.10\%} \\ 
\midrule
        w/ \trc{} & 89.41 $\pm$ 0.21 & 98.28 $\pm$ 0.03 & 90.36 $\pm$ 0.49 & 97.91 $\pm$ 0.01 & 94.24 $\pm$ 0.08 & 87.45 $\pm$ 0.01 & 98.87 $\pm$ 0.13  \\
        Imp. (\%)  & {\color{darkgreen}1.44\%} & {\color{darkgreen}0.22\%} & {\color{darkgreen}0.56\%} & {\color{darkgreen}0.12\%} & {\color{darkgreen}0.04\%} & {\color{darkgreen}0.24\%} & {\color{darkgreen}0.04\%} \\ 
\midrule
        w/ \n{} & {88.97} $\pm$ 0.41 & {98.30} $\pm$ 0.00 & {89.87} $\pm$ 0.06 & {97.85} $\pm$ 0.02 & {95.08} $\pm$ 0.13 & 88.08 $\pm$ 0.23 & {98.91} $\pm$ 0.00 \\
        \multicolumn{1}{r}{Imp. (\%)} & {\color{darkgreen}0.94\%} & {\color{darkgreen}0.24\%} & {\color{darkgreen}0.01\%} & {\color{darkgreen}0.06\%} & {\color{darkgreen}0.93\%} & {\color{darkgreen}0.96\%} & {\color{darkgreen}0.08\%} \\
\bottomrule
\end{tabular}

\vskip 0.1in

\begin{tabular}{c|ccccccc}
\toprule
        \textbf{Methods} & \textbf{SE} & \textbf{UC} & \textbf{UT} & \textbf{UV} & \textbf{US} & \textbf{WK} \\
\midrule
        GraphMixer & 91.73 $\pm$ 0.13 & 91.39 $\pm$ 0.05 & 62.26 $\pm$ 0.13 & 50.58 $\pm$ 0.45 & 50.72 $\pm$ 0.64 & 96.47 $\pm$ 0.13 \\
        w/ \lte{}   & 91.81 $\pm$ 0.11 & 92.07 $\pm$ 0.14 & 61.92 $\pm$ 0.37 & 50.76 $\pm$ 0.34 & 50.87 $\pm$ 0.21 & 96.58 $\pm$ 0.13 \\ 
        Imp. (\%)  & {\color{darkgreen}0.09\%}  & {\color{darkgreen}0.75\%} & {\color{RawSienna}-0.54\%} & {\color{darkgreen}0.37\%} & {\color{darkgreen}0.30\%} & {\color{darkgreen}0.11\%} \\
\midrule
        w/ \trc{} & 91.91 $\pm$ 0.01 & 92.84 $\pm$ 0.13 & 63.05 $\pm$ 0.83 & 55.37 $\pm$ 0.22 & 50.99 $\pm$ 0.96 & 98.11 $\pm$ 0.01 \\
        Imp. (\%) & {\color{darkgreen}0.20\%} & {\color{darkgreen}1.59\%} & {\color{darkgreen}1.28\%} & {\color{darkgreen}9.47\%} & {\color{darkgreen}0.53\%} & {\color{darkgreen}1.70\%} \\
\midrule
        w/ \n{} & 93.03 $\pm$ 0.46 & 93.44 $\pm$ 0.29 & 62.40 $\pm$ 1.50 & 56.63 $\pm$ 0.36 & 51.44 $\pm$ 0.75 & 98.17 $\pm$ 0.01 \\ 
        Imp. (\%)  & {\color{darkgreen}1.44\%} & {\color{darkgreen}2.24\%} & {\color{darkgreen}0.22\%} & {\color{darkgreen}11.96\%} & {\color{darkgreen}1.42\%} & {\color{darkgreen}1.76\%} \\
\midrule
\noalign{\vskip 0.05in}
\midrule
        DyGFormer & 93.16 $\pm$ 0.08 & 94.40 $\pm$ 0.21 & 63.86 $\pm$ 0.35 & 56.16 $\pm$ 0.07 & 54.93 $\pm$ 0.53 & 98.54 $\pm$ 0.07 \\
        w/ \lte{} & 93.21 $\pm$ 0.10  & 95.62 $\pm$ 0.05 & 64.94 $\pm$ 0.08 & 56.27 $\pm$ 0.16 & 54.06 $\pm$ 1.03 & 98.86 $\pm$ 0.01 \\
        Imp. (\%) & {\color{darkgreen}0.05\%} & {\color{darkgreen}1.29\%} & {\color{darkgreen}1.69\%} & {\color{darkgreen}0.20\%} & {\color{RawSienna}-1.58\%} & {\color{darkgreen}0.32\%} \\
\midrule
        w/ \trc{} & 94.49 $\pm$ 0.01 & 94.62 $\pm$ 0.01 & {63.65} $\pm$ 0.18 & {56.20} $\pm$ 0.47 & {55.85} $\pm$ 0.20 & 98.63 $\pm$ 0.06 \\
        Imp. (\%)  & {\color{darkgreen}1.42\%} & {\color{darkgreen}0.23\%} & {\color{RawSienna}-0.33\%} & {\color{darkgreen}0.07\%} & {\color{darkgreen}1.67\%} & {\color{darkgreen}0.09\%} \\
        
\midrule
        {DyGFormer}  & {93.17} $\pm$ 0.01 & {95.67} $\pm$ 0.09 & 64.25 $\pm$ 0.24 & 56.23 $\pm$ 0.17 & {55.08} $\pm$ 0.21 & {98.88}$\pm$ 0.07 \\
        
        \multicolumn{1}{r}{Imp. (\%)} & {\color{darkgreen}0.01\%} & {\color{darkgreen}1.35\%} & {\color{darkgreen}0.61\%} & {\color{darkgreen}0.12\%} & {\color{darkgreen}0.27\%} & {\color{darkgreen}0.35\%} \\
\bottomrule
\end{tabular}
\end{table*}


\setlength{\tabcolsep}{2pt}
\begin{table*}[t!]
\caption{Inductive link prediction performance. AP of methods under various numbers of negative links during testing. \textbf{NEG=50} indicates that each positive link is evaluated against 50 negative links in the AP computation. For dataset names, please kindly refer to Section~\ref{appendix:dataset_description}.}
\label{tab:appendix_inductive_more_negative}
\centering
\tiny

\begin{tabular}{c|cccc||c|cccc}
\toprule
      \textbf{Methods} & \textbf{EN} & \textbf{LA} & \textbf{UC} & \textbf{UV}  & \textbf{Methods} & \textbf{EN} & \textbf{LA} & \textbf{UC} & \textbf{UV} \\
\midrule
        \multicolumn{1}{l}{\textbf{NEG = 1}} & \\
\midrule
        GraphMixer  & 75.94 $\pm$ 0.21 & 82.06 $\pm$ 0.25 & 91.39 $\pm$ 0.05 & 50.58 $\pm$ 0.45 & DyGFormer  & 89.86 $\pm$ 0.28 & 94.20 $\pm$ 0.17 & 94.40 $\pm$ 0.21 & 56.16 $\pm$ 0.07 \\
        w/ \n{}  & 85.54 $\pm$ 0.45 & 90.71 $\pm$ 0.64 & 93.44 $\pm$ 0.29 & 56.63 $\pm$ 0.36 & w/ \n{}  & 89.87 $\pm$ 0.06 & 95.08 $\pm$ 0.13 & 95.67 $\pm$ 0.09 & 56.23 $\pm$ 0.17 \\
        Imp. (\%)  & {\color{darkgreen}12.63\%} & {\color{darkgreen}10.55\%} & {\color{darkgreen}2.24\%} & {\color{darkgreen}11.96\%} & Imp. (\%) &  {\color{darkgreen}0.01\%} & {\color{darkgreen}0.93\%} & {\color{darkgreen}1.35\%} & {\color{darkgreen}0.12\%} \\
\midrule
        \multicolumn{1}{l}{\textbf{NEG = 5}} & \\
\midrule
        GraphMixer & 41.80 $\pm$ 0.35 & 58.37 $\pm$ 0.69 & 77.20 $\pm$ 0.05 & 17.38 $\pm$ 0.62 & DyGFormer  & 66.88 $\pm$ 0.33 & 81.90 $\pm$ 0.52 & 85.86 $\pm$ 0.45 & 21.26 $\pm$ 0.30 \\
        w/ \n{} & 57.22 $\pm$ 0.37 & 72.63 $\pm$ 1.95 & 81.75 $\pm$ 0.74 & 20.89 $\pm$ 0.21 & w/ \n{} & 67.31 $\pm$ 0.08 & 84.37 $\pm$ 0.59 & 88.02 $\pm$ 0.28 & 21.50 $\pm$ 0.47 \\
        Imp. (\%)  & {\color{darkgreen}36.91\%} & {\color{darkgreen}24.43\%} & {\color{darkgreen}5.90\%} & {\color{darkgreen}20.20\%} & Imp. (\%)  & {\color{darkgreen}0.64\%} & {\color{darkgreen}3.01\%} & {\color{darkgreen}2.51\%} & {\color{darkgreen}1.13\%} \\
\midrule
        \multicolumn{1}{l}{\textbf{NEG = 25}} & \\
\midrule
        GraphMixer  & 15.16 $\pm$ 0.33 & 33.93 $\pm$ 1.25 & 55.31 $\pm$ 0.21 & 4.14 $\pm$ 0.22 & DyGFormer  & 34.66 $\pm$ 0.24 & 58.96 $\pm$ 1.15 & 71.89 $\pm$ 1.73 & 5.31 $\pm$ 0.13\\
        w/ \n{}  & 24.19 $\pm$ 0.09 & 45.18 $\pm$ 4.36 & 60.48 $\pm$ 1.58 & 5.09 $\pm$ 0.08 & w/ \n{}  & 35.85 $\pm$ 0.04 & 63.78 $\pm$ 2.10 & 75.52 $\pm$ 0.51 & 5.43 $\pm$ 0.11 \\
        Imp. (\%)  & {\color{darkgreen}59.58\%} & {\color{darkgreen}33.18\%} & {\color{darkgreen}9.35\%} & {\color{darkgreen}22.97\%} & Imp. (\%)  & {\color{darkgreen}3.42\%} & {\color{darkgreen}8.18\%} & {\color{darkgreen}5.06\%} & {\color{darkgreen}2.26\%} \\
\midrule
        \multicolumn{1}{l}{\textbf{NEG = 50}} & \\
\midrule
        GraphMixer  & 8.87 $\pm$ 0.21 & 24.85 $\pm$ 1.30 & 45.00 $\pm$ 0.51 & 2.15 $\pm$ 0.15 & DyGFormer  & 22.53 $\pm$ 0.04 & 46.74 $\pm$ 1.36 & 62.61 $\pm$ 3.26 & 2.76 $\pm$ 0.08\\
        w/ \n{}  & 14.50 $\pm$ 0.30 & 33.66 $\pm$ 4.48 & 49.54 $\pm$ 2.18 & 2.64 $\pm$ 0.05 & w/ \n{} & 23.72 $\pm$ 0.06 & 53.47 $\pm$ 1.61 & 67.75 $\pm$ 0.70 & 2.79 $\pm$ 0.05 \\
        Imp. (\%)  & {\color{darkgreen}63.42\%} & {\color{darkgreen}35.45\%} & {\color{darkgreen}10.08\%} & {\color{darkgreen}22.84\%} & Imp. (\%) & {\color{darkgreen}5.28\%} & {\color{darkgreen}14.39\%} & {\color{darkgreen}8.21\%} & {\color{darkgreen}1.09\%} \\
\bottomrule
\end{tabular}

\end{table*}
\setlength{\tabcolsep}{4pt}

\begin{table*}[h!]
\caption{AP for inductive link prediction. The proposed \lte{} and \trc{} can be integrated into various types of TGNN. No \lte{} and \n{} results are reported for JODIE because it does not utilize the time encoding function.}
\label{tab:appendix_inductive_portability}
\centering
\scriptsize

\begin{tabular}{ccccc}
\toprule
        \textbf{Method} & \textbf{EN} & \textbf{LA} & \textbf{UC} & \textbf{UV} \\
\midrule
        TGN  & 78.74 $\pm$ 1.53 & 80.05 $\pm$ 4.00 & 86.66 $\pm$ 1.83 & 54.13 $\pm$ 2.37 \\
        TGAT & 67.63 $\pm$ 0.96 & 78.48 $\pm$ 0.07 & 79.11 $\pm$ 0.28 & 51.98 $\pm$ 0.14 \\
        CAWN & 84.97 $\pm$ 0.21 & 89.48 $\pm$ 0.02 & 92.45 $\pm$ 0.06 & 49.29 $\pm$ 0.87 \\
        JODIE  & 80.67 $\pm$ 2.11 & 78.31 $\pm$ 3.82 & 79.27 $\pm$ 1.97 & 56.86 $\pm$ 0.06 \\
        
\midrule
        \multicolumn{5}{l}{\textbf{w/ \lte{}}} \\
\midrule
        
        TGN  & 84.02 $\pm$ 1.63 & 89.33 $\pm$ 0.87 & 91.56 $\pm$ 1.41 & 57.92 $\pm$ 3.41 \\
        \multicolumn{1}{r}{Imp. (\%)}  & {\color{darkgreen}6.71\%} & {\color{darkgreen}11.59\%} & {\color{darkgreen}5.66\%} & {\color{darkgreen}7.01\%} \\
\midrule
        
        TGAT  & 77.70 $\pm$ 0.34 & 88.27 $\pm$ 0.05 & 92.61 $\pm$ 0.29 & 50.03 $\pm$ 0.31 \\
        \multicolumn{1}{r}{Imp. (\%)}  & {\color{darkgreen}14.89\%} & {\color{darkgreen}12.47\%} & {\color{darkgreen}17.07\%} & {\color{RawSienna}-3.75\%} \\
\midrule
        
        CAWN & 86.88 $\pm$ 0.13 & 91.14 $\pm$ 0.05 & 94.64 $\pm$ 0.08 & 48.54 $\pm$ 0.25 \\
        \multicolumn{1}{r}{Imp. (\%)}   & {\color{darkgreen}2.25\%} & {\color{darkgreen}1.86\%} & {\color{darkgreen}2.37\%} & {\color{RawSienna}-1.51\%} \\
        
\midrule
        \multicolumn{5}{l}{\textbf{w/ \trc{}}} \\
\midrule
        
        TGN  & 85.97 $\pm$ 0.78 & 92.35 $\pm$ 0.44 & 89.55 $\pm$ 0.18 & 57.26 $\pm$ 3.58 \\
        \multicolumn{1}{r}{Imp. (\%)}  & {\color{darkgreen}9.18\%} & {\color{darkgreen}15.37\%} & {\color{darkgreen}3.34\%} & {\color{darkgreen}5.79\%} \\
\midrule
        
        TGAT  & 83.89 $\pm$ 0.01 & 90.26 $\pm$ 0.33 & 88.16 $\pm$ 0.13 & 54.42 $\pm$ 1.02 \\
        \multicolumn{1}{r}{Imp. (\%)}  & {\color{darkgreen}24.04\%} & {\color{darkgreen}15.01\%} & {\color{darkgreen}11.44\%} & {\color{darkgreen}4.69\%} \\
\midrule
        
        CAWN  & 86.24 $\pm$ 0.15 & 91.03 $\pm$ 0.06 & 92.74 $\pm$ 0.06 & 55.62 $\pm$ 1.66 \\
        \multicolumn{1}{r}{Imp. (\%)}   & {\color{darkgreen}1.49\%} & {\color{darkgreen}1.73\%} & {\color{darkgreen}0.31\%} & {\color{darkgreen}12.84\%} \\
\midrule
        
        JODIE  & 85.48 $\pm$ 0.24 & 91.71 $\pm$ 0.56 & 83.56 $\pm$ 0.08 & 57.64 $\pm$ 1.41 \\
        \multicolumn{1}{r}{Imp. (\%)}  & {\color{darkgreen}5.97\%} & {\color{darkgreen}17.11\%} & {\color{darkgreen}5.42\%} & {\color{darkgreen}1.37\%} \\

\midrule
        \multicolumn{5}{l}{\textbf{w/ \n{}}} \\
\midrule

        TGN   & 85.54 $\pm$ 0.45 & 90.71 $\pm$ 0.64 & 93.44 $\pm$ 0.29 & 57.63 $\pm$ 0.36 \\
        \multicolumn{1}{r}{Imp. (\%)}   & {\color{darkgreen}8.64\%} & {\color{darkgreen}13.32\%} & {\color{darkgreen}7.82\%} & {\color{darkgreen}6.47\%} \\
\midrule
        
        TGAT   & 84.53 $\pm$ 0.13 & 93.32 $\pm$ 0.20 & 94.25 $\pm$ 0.10 & 54.29 $\pm$ 0.30 \\
        \multicolumn{1}{r}{Imp. (\%)}  & {\color{darkgreen}24.98\%} & {\color{darkgreen}18.91\%} & {\color{darkgreen}19.15\%} & {\color{darkgreen}4.44\%} \\
\midrule
        
        CAWN  & 87.45 $\pm$ 0.23 & 92.61 $\pm$ 0.04 & 94.69 $\pm$ 0.11 & 64.33 $\pm$ 1.37 \\
        \multicolumn{1}{r}{Imp. (\%)}   & {\color{darkgreen}2.92\%} & {\color{darkgreen}3.50\%} & {\color{darkgreen}2.42\%} & {\color{darkgreen}30.53\%} \\
\midrule
    \end{tabular}

\end{table*}

\section{Experimental Settings on the TGB Datasets}
\label{appendix_sec_tgb_exp_settings}
We further evaluate \n{} on three large-scale datasets from the Temporal Graph Benchmark (TGB)~\cite{tgb-dataset}, including tgbl-wiki-v2, tgbl-review-v2, and tgbl-coin-v2. Dataset statistics are available at \href{https://tgb.complexdatalab.com/docs/linkprop/}{https://tgb.complexdatalab.com/docs/linkprop/}. We apply \n{} to the recent state-of-the-art method DyGFormer and compare its performance against the original version. For fair comparison, we adopt the best hyperparameter settings of DyGFormer on the TGB benchmark, as detailed in~\cite{dyg_tgb_setting}. In our \trc{} module, the hyperparameter $\gamma$ is fixed at 0.9 across the training, validation, and testing phases. The dimension of historical edge embeddings is equal to the dimension of temporal node embeddings. The detailed results are presented in Table~\ref{tab:tgb_results}. We also report the original skewness of temporal differences in the TGB datasets and the skewness after applying our LTE in Table~\ref{tab:appendix_tgb_skewness} below. Results show that our LTE effectively reduces the skewness in the distribution of the temporal differences.

\begin{table}[t]
    \centering
    \caption{Skewness of temporal differences before and after applying our LTE on the three TGB datasets tested.}
    \begin{tabular}{c|ccc}
    \toprule
    \textbf{Datasets} & \textbf{tgbl-wiki} & \textbf{tgbl-review} & \textbf{tgbl-coin} \\
    \midrule
    Originally & 4.181 & 3.297 & 4.562 \\
    \midrule
    with LTE &  -0.307 & -0.524 & 0.308 \\
    \bottomrule
    \end{tabular}
    \label{tab:appendix_tgb_skewness}
\end{table}


\section{Standard Deviations in Main Text}
\label{sec:std_main_text}
Table~\ref{tab:std_main_table}, Table~\ref{tab:std_ablation}, and Table~\ref{tab:std_more_negative} present the standard deviations of the results shown in Table~\ref{tab:main_result}, Table~\ref{tab:ablation}, and Table~\ref{tab:exp_more_negative} in the main text, respectively.
\setlength{\tabcolsep}{5pt}
\begin{table*}[h!]
\caption{The standard deviations of five runs for results in Table~\ref{tab:main_result} and Table~\ref{tab:appendix_full_main_results}.}
\label{tab:std_main_table}
\centering
\tiny

\begin{tabular}{c|c|ccccccccccccc}
\toprule

\textbf{NSS} & \textbf{Methods} & \textbf{CP} & \textbf{CO} & \textbf{EN} & \textbf{FL} & \textbf{LA} & \textbf{MO} & \textbf{RE} & \textbf{SE} & \textbf{UC} & \textbf{UT} & \textbf{UV} & \textbf{US} & \textbf{WK} \\

\midrule
\multirow{14}{*}{rnd} & JODIE & 0.31 & 1.33 & 0.30 & 1.73 & 2.13 & 2.44 & 0.14 & 0.55 & 1.09 & 0.31 & 0.81 & 1.52 & 0.14 \\
        & DyRep & 2.76 & 0.15 & 3.36 & 0.72 & 2.21 & 0.49 & 0.04 & 0.30 & 2.30 & 0.93 & 0.80 & 0.39 & 0.06 \\
        & TGAT & 0.72 & 0.09 & 0.97 & 0.18 & 0.21 & 0.15 & 0.02 & 0.17 & 0.70 & 0.18 & 0.98 & 3.16 & 0.06 \\
        & TGN & 2.34 & 0.56 & 1.11 & 0.14 & 3.97 & 1.60 & 0.06 & 0.17 & 1.04 & 1.37 & 2.17 & 0.58 & 0.06 \\
        & CAWN & 2.34 & 0.28 & 0.09 & 0.01 & 0.06 & 0.25 & 0.01 & 0.09 & 0.06 & 0.12 & 0.10 & 0.48 & 0.03 \\
        & EdgeBank & 0.00 & 0.00 & 0.00 & 0.00 & 0.00 & 0.00 & 0.00 & 0.00 & 0.00 & 0.00 & 0.00 & 0.00 & 0.00 \\
        & TCL & 2.67 & 0.12 & 0.71 & 0.02 & 2.16 & 0.24 & 0.02 & 0.16 & 1.63 & 0.03 & 0.03 & 0.48 & 0.16 \\
        & GraphMixer & 0.46 & 0.09 & 0.06 & 0.01 & 0.02 & 0.04 & 0.01 & 0.16 & 0.05 & 0.13 & 0.09 & 0.23 & 0.04 \\
        & DyGFormer & 1.89 & 0.08 & 0.50 & 0.01 & 1.53 & 0.17 & 0.01 & 0.11 & 1.15 & 0.02 & 0.02 & 0.34 & 0.11 \\
\cmidrule{2-15}
& \multicolumn{13}{l}{\textbf{w/ \n{}}} \\
\cmidrule{2-15}
        & GraphMixer & 0.07 & 0.06 & 0.06 & 0.01 & 0.41 & 0.04 & 0.04 & 0.11 & 0.21 & 0.49 & 0.08 & 0.36 & 0.00 \\
        & DyGFormer & 0.12 & 0.01 & 0.21 & 0.00 & 0.08 & 0.14 & 0.00 & 0.03 & 0.21 & 0.04 & 0.03 & 0.28 & 0.01 \\

\midrule

\multirow{14}{*}{hist} & JODIE & 0.63 & 2.13 & 2.70 & 2.59 & 3.81 & 1.25 & 0.36 & 6.78 & 5.80 & 1.83 & 0.81 & 5.76 & 0.66 \\
& DyRep & 1.23 & 0.20 & 2.76 & 0.99 & 2.45 & 1.12 & 0.31 & 0.43 & 3.14 & 1.07 & 1.12 & 2.25 & 0.56 \\
& TGAT & 0.84 & 0.52 & 1.05 & 0.18 & 0.24 & 0.62 & 0.20 & 0.44 & 1.37 & 0.91 & 2.14 & 6.60 & 0.22 \\
& TGN & 3.07 & 2.35 & 1.76 & 1.64 & 4.64 & 1.93 & 0.61 & 0.56 & 2.12 & 5.51 & 3.93 & 7.57 & 0.33 \\
& CAWN & 2.77 & 0.49 & 0.36 & 0.97 & 0.43 & 0.95 & 0.45 & 0.38 & 0.43 & 0.38 & 0.04 & 8.23 & 1.67 \\
& EdgeBank & 0.00 & 0.00 & 0.00 & 0.00 & 0.00 & 0.00 & 0.00 & 0.00 & 0.00 & 0.00 & 0.00 & 0.00 & 0.00 \\
& TCL & 3.00 & 0.21 & 0.39 & 0.24 & 2.31 & 0.41 & 0.16 & 0.31 & 2.74 & 1.17 & 2.35 & 3.95 & 0.39 \\
& GraphMixer & 0.87 & 0.41 & 0.92 & 0.26 & 0.49 & 0.92 & 0.18 & 0.31 & 1.35 & 1.22 & 1.60 & 1.02 & 0.10 \\
& DyGFormer & 0.31 & 0.06 & 0.23 & 0.49 & 0.48 & 0.66 & 0.67 & 0.14 & 0.82 & 1.40 & 1.58 & 3.38 & 2.56 \\

\cmidrule{2-15}
& \multicolumn{13}{l}{\textbf{w/ \n{}}} \\
\cmidrule{2-15}

& GraphMixer & 0.12 & 0.02 & 0.68 & 0.11 & 1.06 & 0.62 & 0.29 & 0.03 & 0.53 & 2.15 & 0.32 & 1.11 & 0.57 \\

& DyGFormer & 0.01 & 0.02 & 0.01 & 2.93 & 1.84 & 0.89 & 0.10 & 0.04 & 0.32 & 1.46 & 2.12 & 2.50 & 0.24 \\

\midrule

\multirow{14}{*}{ind} & JODIE     & 0.66 & 1.78 & 0.98 & 4.02 & 4.49 & 2.19 & 0.16 & 4.11 & 1.40 & 1.48 & 1.48 & 5.13 & 0.79 \\
& DyRep     & 0.86 & 0.10 & 1.53 & 1.82 & 2.70 & 0.95 & 0.26 & 0.48 & 1.76 & 1.24 & 1.24 & 1.16 & 1.58 \\
& TGAT      & 1.21 & 0.48 & 1.36 & 0.26 & 0.17 & 0.64 & 0.24 & 0.44 & 0.84 & 1.24 & 1.24 & 5.82 & 0.16 \\
& TGN       & 2.87 & 3.28 & 2.72 & 2.72 & 5.98 & 2.91 & 0.24 & 0.56 & 0.71 & 6.01 & 6.01 & 6.47 & 0.44 \\
& CAWN      & 1.00 & 0.27 & 0.58 & 1.52 & 0.77 & 0.94 & 0.24 & 0.27 & 0.48 & 0.67 & 0.67 & 8.52 & 2.62 \\
& EdgeBank  & 0.00 & 0.00 & 0.00 & 0.00 & 0.00 & 0.00 & 0.00 & 0.00 & 0.00 & 0.00 & 0.00 & 0.00 & 0.00 \\
& TCL       & 1.75 & 0.21 & 0.32 & 0.18 & 0.89 & 0.54 & 0.29 & 0.36 & 1.11 & 1.74 & 1.74 & 3.34 & 0.72 \\
& GraphMixer& 0.63 & 0.35 & 0.79 & 0.21 & 0.33 & 0.92 & 0.11 & 0.33 & 0.51 & 1.29 & 1.29 & 0.84 & 0.17 \\
& DyGFormer & 0.57 & 0.28 & 0.89 & 1.78 & 0.50 & 0.69 & 0.40 & 0.10 & 1.71 & 1.02 & 1.02 & 3.62 & 5.38 \\
\cmidrule{2-15}
& \multicolumn{13}{l}{\textbf{w/ \n{}}} \\
\cmidrule{2-15}
& GraphMixer & 0.28 & 0.74 & 0.29 & 0.69 & 0.06 & 0.38 & 0.09 & 0.00 & 0.39 & 0.68 & 0.04 & 0.01 & 0.24 \\
& DyGFormer  & 0.09 & 0.06 & 1.39 & 2.82 & 1.63 & 1.38 & 0.21 & 0.21 & 0.35 & 3.27 & 1.44 & 2.86 & 0.08 \\
\bottomrule
\end{tabular}
\vskip -0.2in
\end{table*}

\begin{table*}[h!]
\caption{The standard deviations of five runs for results in Table~\ref{tab:ablation}.}
\label{tab:std_ablation}
\centering
\tiny
\begin{tabular}{lccccccccccccc}
\toprule

\textbf{Methods} & \textbf{CP} & \textbf{CO} & \textbf{EN} & \textbf{FL} & \textbf{LA} & \textbf{MO} & \textbf{RE} & \textbf{SE} & \textbf{UC} & \textbf{UT} & \textbf{UV} & \textbf{US} & \textbf{WK} \\

\midrule
        GraphMixer & 0.46 & 0.09 & 0.06 & 0.01 & 0.02 & 0.04 & 0.01 & 0.16 & 0.05 & 0.13 & 0.09 & 0.23 & 0.04 \\
        w/ \lte{} & 0.11 & 0.18 & 0.34 & 0.01 & 0.45 & 0.15 & 0.01 & 0.12 & 0.02 & 0.35 & 0.23 & 0.18 & 0.01 \\
        w/ \trc{} & 0.60 & 0.05 & 0.04 & 0.03 & 0.07 & 0.19 & 0.02 & 0.06 & 0.18 & 0.19 & 0.17 & 0.33 & 0.00 \\
        w/ \n{} & 0.07 & 0.06 & 0.06 & 0.01 & 0.41 & 0.04 & 0.04 & 0.11 & 0.21 & 0.29 & 0.08 & 0.36 & 0.00 \\
\midrule
        DyGFormer & 0.26 & 0.00 & 0.04 & 0.01 & 0.19 & 0.24 & 0.01 & 0.08 & 0.30 & 0.94 & 0.32 & 0.91 & 0.05 \\
        w/ \lte{} & 0.07 & 0.01 & 0.06 & 0.00 & 0.00 & 0.11 & 0.01 & 0.01 & 0.22 & 0.01 & 0.45 & 0.69 & 0.02 \\
        w/ \trc{} & 0.12 & 0.01 & 0.06 & 0.01 & 0.34 & 0.33 & 0.01 & 0.00 & 0.05 & 0.66 & 1.09 & 0.51 & 0.05 \\
        w/ \n{} & 0.12 & 0.01 & 0.21 & 0.00 & 0.08 & 0.14 & 0.00 & 0.03 & 0.21 & 0.04 & 0.03 & 0.28 & 0.01 \\
\bottomrule
\end{tabular}
\vskip -0.2in
\end{table*}

\begin{table*}[t]
\caption{The standard deviations of five runs for results in Table~\ref{tab:exp_more_negative}.}
\label{tab:std_more_negative}
\centering
\tiny
\begin{tabular}{c|cccc||c|cccc}
\toprule
        \textbf{Method} & \textbf{EN} & \textbf{LA} & \textbf{UC} & \textbf{UV} & \textbf{Method} &  \textbf{EN} & \textbf{LA} & \textbf{UC} & \textbf{UV} \\
\midrule
        \multicolumn{1}{l}{\textbf{NEG = 1}} & \\
\midrule
        GraphMixer   & 0.06 & 0.02 & 0.05 & 0.09 & DyGFormer & 0.04 & 0.19 & 0.30 & 0.32 \\
        w/ \n{}  & 0.06 & 0.41 & 0.21 & 0.08 & w/ \n{}  & 0.21 & 0.08 & 0.21 & 0.33 \\
\midrule
        \multicolumn{1}{l}{\textbf{NEG = 5}} & \\ 
\midrule
        GraphMixer  & 0.39 & 0.23 & 0.09 & 0.16 & DyGFormer  & 0.07 & 0.71 & 0.69 & 0.26 \\
        w/ \n{}  & 0.68 & 1.36 & 0.43 & 0.04 & w/ \n{} & 0.62 & 0.51 & 0.46 & 0.44 \\
\midrule
        \multicolumn{1}{l}{\textbf{NEG = 25}} & \\ 
\midrule
        GraphMixer  & 0.20 & 0.53 & 0.27 & 0.05 & DyGFormer  & 0.04 & 1.46 & 1.18 & 0.19 \\
        w/ \n{}   & 1.19 & 2.62 & 1.10 & 0.02 & w/ \n{}  & 0.52 & 1.58 & 0.57 & 0.13 \\
\midrule
        \multicolumn{1}{l}{\textbf{NEG = 50}} & \\ 
\midrule
        GraphMixer   & 0.16 & 0.52 & 0.08 & 0.04 & DyGFormer   & 0.11 & 1.82 & 1.60 & 0.06 \\
        w/ \n{}  & 0.90 & 2.54 & 1.73 & 0.01 & w/ \n{}   & 0.59 & 2.13 & 0.53 & 0.11 \\
\bottomrule
\end{tabular}
\end{table*}

\end{document}